%% file: neurips_2026.tex
\documentclass{article}

\PassOptionsToPackage{numbers,compress}{natbib}
\usepackage[preprint]{neurips_2026}

\usepackage[utf8]{inputenc}
\usepackage[T1]{fontenc}
\usepackage[colorlinks=true, citecolor=cyan, linkcolor=black]{hyperref}
\usepackage{url}
\usepackage{booktabs}
\usepackage{amsfonts}
\usepackage{amsmath}
\usepackage{nicefrac}
\usepackage{microtype}
\usepackage{xcolor}
\usepackage{graphicx}
\usepackage{enumitem}
\usepackage{xspace}
\usepackage{pifont}
\usepackage{multirow}
\usepackage{colortbl}
\usepackage{makecell}
\usepackage{float}
\usepackage{placeins}
\usepackage{tabularx}
\usepackage{enumitem}
\usepackage{longtable}
\usepackage{array}
\usepackage{makecell}
\usepackage{wrapfig}
\usepackage{tabularx}
\usepackage{booktabs}
\usepackage{array}
\usepackage{makecell}

\usepackage{amsthm}
\newtheorem{proposition}{Proposition}

\newcolumntype{L}[1]{>{\raggedright\arraybackslash}p{#1}}
\newcolumntype{C}[1]{>{\centering\arraybackslash}p{#1}}
\newcolumntype{Y}{>{\raggedright\arraybackslash}X}

\definecolor{ilrow}{HTML}{E8F1FB}
\definecolor{vlarow}{HTML}{E8F5E9}
\definecolor{wamrow}{HTML}{FFF3E0}

\newcommand{\titleicon}{%
  \raisebox{-0.18em}{\includegraphics[height=1.15em]{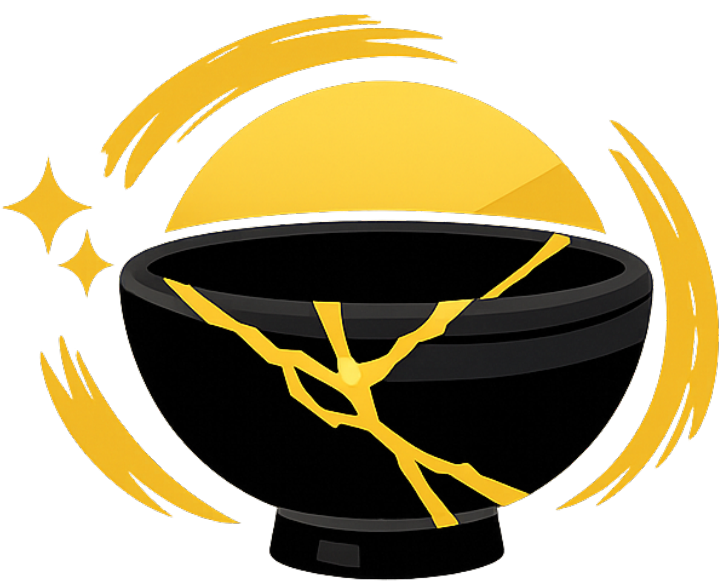}}%
}

\title{\texorpdfstring{\titleicon\hspace{0.25em}Kintsugi:}{Kintsugi:} Learning Policies by  Repairing Executable Knowledge Bases}

\author{%
\begin{tabular}{c}
Teng Cao\textsuperscript{1,2,3,*}
\quad
Yu Deng\textsuperscript{1,2,3,*}
\quad
Hikaru Shindo\textsuperscript{1,3,*}
\\
Quentin Delfosse\textsuperscript{1,3\dag}
\quad
Lanxi Wen\textsuperscript{4}
\quad
Suli Wang\textsuperscript{3}
\\
Jannis Blüml\textsuperscript{1,3}
\quad
Christopher Tauchmann\textsuperscript{1,2,3}
\quad
Kristian Kersting\textsuperscript{1,3,2,5,6}
\end{tabular}
\\[0.8em]
\parbox{0.95\textwidth}{\centering\small
\textsuperscript{1}Artificial Intelligence and Machine Learning Lab, Technical University of Darmstadt, Germany
\\
\textsuperscript{2}Hessian Center for Artificial Intelligence (hessian.AI), Germany
\\
\textsuperscript{3}Department of Computer Science, Technical University of Darmstadt, Germany
\\
\textsuperscript{4}Department of Computer Science, Technical University of Munich (TUM), Germany
\\
\textsuperscript{5}German Research Center for Artificial Intelligence (DFKI), Germany
\\
\textsuperscript{6}Centre for Cognitive Science, Technical University of Darmstadt, Germany
\\[0.4em]
\textsuperscript{*}Equal contribution.
\quad
\textsuperscript{\dag}Work done while at the Artificial Intelligence and Machine Learning Lab of the Technical University of Darmstadt; now at Intrinsic AI, Google.
\\[0.4em]
\texttt{\{teng.cao,yu.deng,kersting\}@tu-darmstadt.de}
}
}

\begin{document}

\maketitle

\input{sections/00_abstract.tex}

\input{sections/01_introduction.tex}
\input{sections/Background}
\input{sections/03_method.tex}
\input{sections/04_experiments.tex}
\input{sections/02_related_work.tex}
\input{sections/06_conclusion.tex}


\bibliographystyle{plainnat}
\bibliography{references}

\clearpage
\appendix
\input{sections/A_appendix.tex}

\newpage

\end{document}

%% file: sections/00_abstract.tex
\begin{abstract}
Modern embodied agents achieve impressive performance, but their task knowledge is often stored in neural weights, latent state, or prompt-bound memory, making individual policy knowledge difficult to inspect, validate, recombine, and reuse. We introduce \textbf{Kintsugi}, a white-box policy-learning framework that treats embodied policy improvement as verifier-gated construction of a typed executable Knowledge Base (KB). Kintsugi represents task-level policy knowledge as composable typed entries---predicates, operators, policy schemas, monitors, recovery rules, experience records, and goals---and improves this artifact through localized typed edits induced from rollout evidence, rather than relying on test-time language-model reasoning. Between rollouts, a tool-constrained agentic editing loop diagnoses trajectory failures, localizes them to editable KB layers, and proposes candidate edits. A deterministic verification gate admits an edit only when the candidate type-checks, the resulting KB executes, and focused validation success or trajectory-health metrics improve without violating protected-regression checks. At inference, the accepted KB is executed by a deterministic symbolic executor with zero LLM calls. Across long-horizon text-agent benchmarks and representative object-centric manipulation settings, Kintsugi achieves strong endpoint performance while preserving inspectability, local editability, and verifier-gated deployment. These results suggest that embodied policy improvement can be organized around executable task knowledge.
\end{abstract}

%% file: sections/01_introduction.tex
\section{Introduction}
A deployed embodied agent must not only act; it must be maintained. When an agent repeatedly opens the wrong container, misses a precondition, or applies a skill to the wrong object, the central question is not how to resample the next action, but how to repair the task knowledge that made the behavior plausible. Recent embodied policies have made rapid progress by scaling data, models, and action-generation architectures. Vision-Language-Action models ~\cite{kim2024openvla, team2024octo, black2024pi_0} and diffusion-style policies can acquire broad perceptual and motor competence~\cite{chi2025diffusion, deng2026robot}, while world-model methods learn compact latent dynamics for planning and control~\cite{hafner2023mastering}. We view these systems as complementary to our setting: learned perception and motor skills may provide the substrate, but their task knowledge is often stored in parameters, latent state, or prompt-bound memory~\cite{zitkovich2023rt, shinn2023reflexion}. In such systems, a repeated deployment failure does not naturally expose a stable edit locus. Failure evidence therefore often becomes a training example, a fine-tuning datum, or an engineer's debugging note, rather than a durable correction to explicit policy knowledge~\cite{wang2024agent, chen2024automanual}.

Symbolic planning and neuro-symbolic abstraction learning address the complementary side of this problem because they expose predicates, preconditions, effects, and plans as inspectable structures~\cite{kaelbling2011hierarchical, garrett2020pddlstream, konidaris2018skillstosymbols}. Classical systems, however, are often hand-engineered and difficult to grow from deployment evidence, while recent learned variants commonly target individual components such as predicate invention, operator learning, continuous-parameter sampling, or domain-specific abstract world models~\cite{liang2026exopredicator, li2025visualpredicator, han2024interpret, chitnis2022learning}. LLM-based agents provide another route because they can reflect on failures, generate code, or synthesize plans~\cite{zhao2024expel, liang2023codeaspolicies}, but their conclusions remain prompt-bound or free-form unless extracted into typed executable artifacts and empirically validated~\cite{packer2023memgpt, sumers2023cognitive}. The gap we target is therefore not ``neural versus symbolic,'' but the absence of a policy-maintenance object that turns rollout evidence into typed, executable, and verifier-gated changes.

\begin{figure*}[t]
\centering
\includegraphics[width=\textwidth]{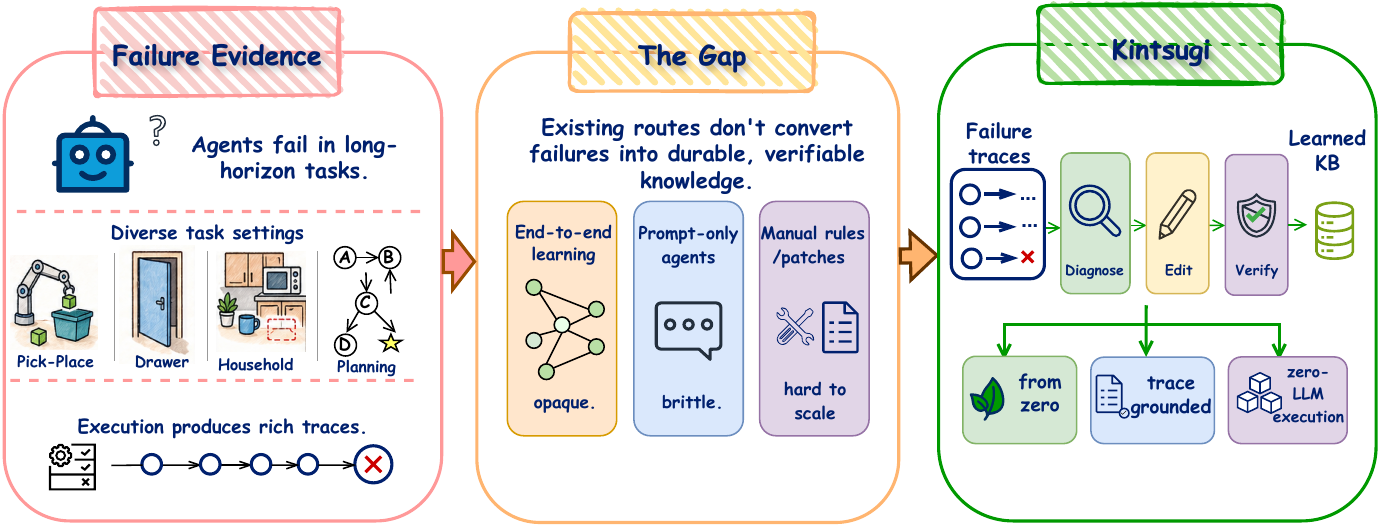}
\caption{\textbf{From failure traces to verified policy knowledge.} Embodied-agent failures provide rich evidence, but end-to-end learning, prompt-only agents, and manual patches often do not yield durable policy repairs. Kintsugi closes this gap by diagnosing failures, editing a typed KB, and verifier-gating updates into traceable policy knowledge that executes with zero LLM calls at inference.}
\label{fig:overview}
\end{figure*}

We study \emph{white-box improvement of the task-knowledge layer} of object-centric embodied policies. The input is trajectory evidence over structured object-centric states and calibrated grounding or skill contracts; the goal is to improve the task policy while keeping decisions and updates inspectable~\cite{konidaris2018skillstosymbols, ahn2022can}. We call this layer white-box when executable decisions can be traced to typed knowledge entries, and updates are localized, rollbackable edits with evidence and validation records. This separation is a deliberate design stance: by decoupling task knowledge from perception and low-level control, the problem formulation remains agnostic to the learned vision models, state estimators, skill libraries, or task-specific motor policies that provide the substrate. The task knowledge above them is represented as an explicit artifact that can be inspected, edited, replayed, and empirically checked.

To instantiate this idea, we propose \textbf{Kintsugi}, a policy-as-editable-knowledge training loop for embodied agents. The name reflects the core design principle: failures are not hidden, but become visible traces for auditable edits to the executable policy artifact. Kintsugi represents task-level policy knowledge as a typed declarative knowledge base (KB) containing grounding rules, predicates, ontology and affordances, operators, policy schemas, monitors and recovery rules, experience records, and goal specifications. Between rollouts, a restricted LLM editor reads trajectory evidence and source contracts, hypothesizes a likely task-knowledge failure layer, and proposes a typed edit. The edit is not accepted because the LLM proposed it: a deterministic applier must type-check the payload, and the resulting KB must pass focused validation and protected-regression checks, which ensure that the edit does not degrade previously solved or explicitly protected behaviors. At inference, Kintsugi makes zero LLM calls. The accepted KB drives the high-level task policy through a deterministic symbolic executor and calibrated skill or action bindings.

This design tests artifact-centered policy improvement rather than claiming that symbolic policies replace neural robotics. We evaluate whether capability is acquired through KB evolution, whether it resides in the executable artifact, whether behavior can be locally edited, and which KB components are load-bearing. On long-horizon text-agent tasks, Kintsugi evaluates endpoint capability on ALFWorld~\cite{shridhar2020alfworld}, WebShop~\cite{yao2022webshop}, and TextCraft~\cite{prasad2024adapt}. On ALFWorld, we further isolate the artifact claim through a cold-start study, fixed-KB access controls, local editability tests, and component ablations. For object-centric manipulation, MetaWorld~\cite{yu2020meta} and Predicators~\cite{silver2023predicateinvention, silver2021symbolicoperators} evaluate KB-driven symbolic policies under compatible state and skill interfaces, while RoboSuite is used as a boundary and failure-localization stress test with simulator state and task-local low-level adapters. Thus the experiments measure success rate, inspectability, traceability, local editability, verifier-gated update discipline, zero-LLM deployment, and the assumptions under which task-level reuse is valid.

Our contributions are:
\begin{itemize}
    \item \textbf{White-box task-knowledge artifact.}
    We formulate embodied policy improvement as constructing and maintaining a durable policy object: an executable typed KB whose high-level decisions, edits, provenance, and validation records are inspectable and replayable under compatible grounding and skill contracts.

    \item \textbf{Layer-localized typed policy repair.}
    We introduce a verifier-gated training loop in which an offline editor agent diagnoses rollout evidence, localizes candidate failures to editable KB layers, and proposes typed policy edits that are accepted only after deterministic type checking, execution checks, focused validation, and protected-regression checks.

    \item \textbf{Artifact-centered deployment and evaluation.}
    We deploy the accepted KB, rather than test-time language reasoning, as the high-level task policy, and evaluate endpoint capability, cold-start KB growth, fixed-KB access controls, local editability, component ablations, reuse, and boundary cases across text-agent and object-centric manipulation settings.
\end{itemize}

%% file: sections/Background.tex
\section{Background and Problem Formulation}
\label{sec:background}
We study object-centric embodied tasks. At each step, the agent observes a structured state \(s_t=\{(o_i,c_i,p_i,r_i,\ldots)\}_{i=1}^{n_t}\), where objects have categories, attributes, poses or relations, and task-relevant states such as open, locked, reachable, held, or occupied~\cite{konidaris2018skillstosymbols, chitnis2022learning}. Given a goal \(g\), the agent acts through a calibrated skill or action interface. Thus the input for policy improvement is object-centric trajectory evidence \(\tau=(s_0,a_0,s_1,a_1,\ldots,s_T,g)\), together with source contracts that define how states, predicates, and skills are exposed~\cite{liang2023codeaspolicies}.

\textbf{White-box policy.} The desired output is an executable policy. A standard policy maps state and goal to an action, \(a_t=\pi(s_t,g)\). We instead study a white-box policy form in which decisions are grounded in an explicit artifact \(K\): \(a_t=E(K,s_t,g)\), with \(\mathrm{Trace}(a_t)\subseteq K\). Here \(E\) is an executor, and \(\mathrm{Trace}(a_t)\) means each deployed policy action can be attributed to specific, identifiable entries in \(K\), rather than explained only after execution.

\textbf{Problem.}
Given object-centric rollout evidence, improve the executable policy while keeping updates inspectable and safe.
A failure should identify a bounded editable locus \(\ell\), apply a local update
\(K'=\mathrm{Apply}(K,\delta_\ell)\), and accept the resulting artifact only if it satisfies focused acceptance without protected regression.
In our setting, focused acceptance means either higher focused success,
\(M_{\mathrm{focus}}(K')>M_{\mathrm{focus}}(K)\), or unchanged focused success together with improvement in a pre-declared trajectory-health metric,
\(M_{\mathrm{focus}}(K')=M_{\mathrm{focus}}(K)\) and
\(H_{\mathrm{focus}}(K')>H_{\mathrm{focus}}(K)\).
In both cases, protected behavior must not regress:
\(M_{\mathrm{protect}}(K')\geq M_{\mathrm{protect}}(K)\).
Thus, the central challenge is to turn object-centric failure evidence into persistent executable policy knowledge whose decisions can be traced, locally edited, and validated before deployment.

%% file: sections/03_method.tex
\begin{figure*}[t]
\centering
\includegraphics[width=\textwidth]{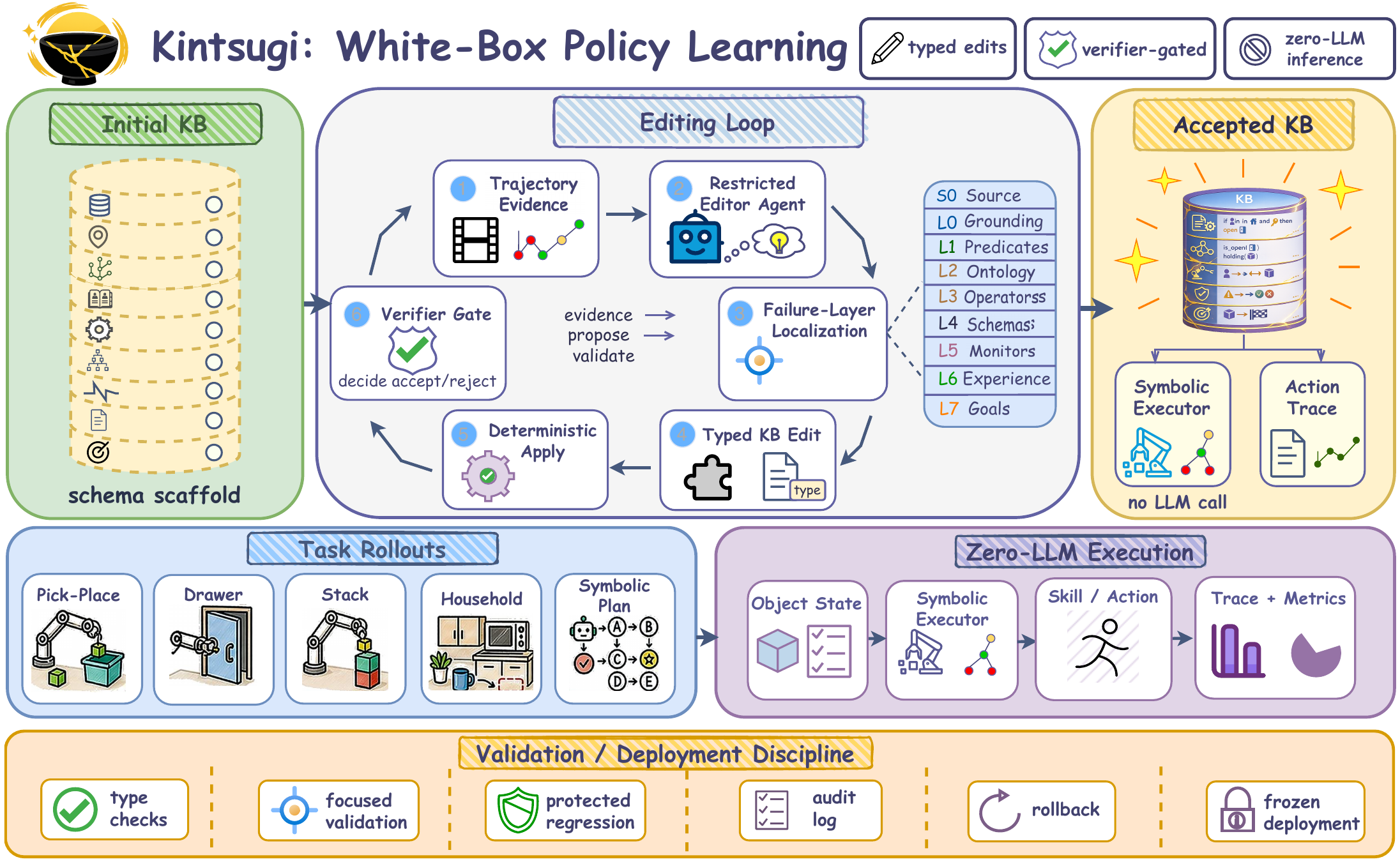}
\caption{\textbf{Verifier-gated KB editing and zero-LLM policy execution in Kintsugi.} During training, a restricted editor uses rollout evidence to localize failures and propose typed KB edits, which are accepted only after deterministic application, focused validation, and protected-regression checks. At inference, the accepted KB drives a separate symbolic-control chain with no LLM calls. Appendix~\ref{app:method-details} details the executable KB schema and KBDiff format.}
\label{fig:workflow}
\end{figure*}

\section{Policy as a Typed Executable Knowledge Base}
\label{sec:kb}

Kintsugi instantiates the policy artifact \(K\) as a typed executable knowledge base (KB). The purpose of the KB is to make task-level policy knowledge explicit: instead of storing a policy only in parameters, hidden states, prompt text, or free-form code, the KB stores the objects, relations, operators, policy schemas, monitors, recovery rules, experience, provenance, and goals that the executor uses to choose actions. Formally, the artifact is organized as typed layers \(K=\{S0,L0,L1,\ldots,L7\}\), where \(S0\) names the observation substrate and \(L0\)--\(L7\) are editable task-knowledge layers. Appendix Table~\ref{tab:app-kb-layers} summarizes the concrete layer vocabulary and representative update loci.

Each KB entry has a typed record form \(e=(\ell,k,\mathrm{type},\mathrm{content},\mathrm{prov})\), where \(\ell\) is the layer, \(k\) is the entry key, \(\mathrm{type}\) specifies the schema, \(\mathrm{content}\) stores the executable knowledge payload, and \(\mathrm{prov}\) records provenance such as trajectory evidence or accepted-edit history. This layer structure is designed so that failures have concrete edit loci. If the agent acts on the wrong object, the likely repair may belong to grounding or ontology. If it misses a required condition, the repair may belong to predicates or operator preconditions. If it chooses a poor skill sequence, the repair may belong to policy schemas. If it fails to detect or recover from a bad state, the repair may belong to monitors or recovery rules. Thus the KB is not only a storage format; it is a structured interface for diagnosis, execution, repair, and long-term policy maintenance.

The entries are typed so that edits can be checked before execution. For example, a predicate entry specifies its argument types and evaluation rule; an operator entry specifies preconditions, effects, and bound skills; a monitor entry specifies its trigger and recovery target; and a goal entry specifies a task decomposition or terminal condition. Accordingly, we write this requirement as \(\mathrm{TypeCheck}(e,\ell)=1\), meaning that entry \(e\) conforms to the schema required by layer \(\ell\). Consequently, a malformed precondition, an operator referring to a nonexistent predicate, or a skill binding with mismatched arguments can be rejected before environment validation.

At inference, the accepted KB is consumed by a deterministic executor, \(a_t=E(K,s_t,g)\). We defer the full execution procedure to Section~\ref{sec:deployment}; here the key point is that KB entries are executable policy components rather than passive memory. This design realizes the white-box requirements from Section~\ref{sec:background}: traceable decisions, localized edits, recorded provenance, and verifier-controlled mutability. Appendix~\ref{app:method-details} gives the concrete artifact layout, KBDiff payload form, and rule-to-skill decision trace used to instantiate these contracts.
\section{Updates as Verifier-Gated Typed Edits}
\label{sec:updates}

A typed policy edit is the only mutation channel for learned policy knowledge. It is not free-form code generation or a reflection string~\cite{liang2023codeaspolicies}. We write an edit as
\[
\Delta = (\mathrm{layer}, \mathrm{key}, \mathrm{op}, \mathrm{payload}, \mathrm{evidence}, \mathrm{metric}, \mathrm{regression\_set}).
\]
The concrete operation vocabulary is benchmark-dependent, but the semantics are shared: create or update an entity, predicate, relation, affordance, operator, policy schema, monitor, recovery rule, skill binding, experience record, or goal schema. In implementation, typed edits are stored as KB diffs with explicit payload schemas; at the paper level, the learned object is the typed policy edit rather than the file patch format. A task-local implementation change may be used as a prototype or adapter repair, but it is not an accepted Kintsugi knowledge update until the learned content is distilled into a typed KB edit and validated by the verifier before deployment.

\paragraph{Example of an accepted edit.}
Consider a failure in which the agent reaches the goal receptacle while holding the target object, but repeatedly attempts to deposit the object into a closed cabinet. The repair is localized to the deposit policy path rather than to navigation or grasping. A valid typed edit targets \(L4\) Policy schemas, key \texttt{rule.OpenGoalRecep}, with operation \texttt{add\_rule}. The payload adds a guard that fires when the task requires deposit, the target object is already held, the agent is at the goal receptacle, and the current receptacle is openable but closed; the corresponding action schema is \texttt{OPEN}. After this rule fires, the existing deposit operator can place the object into the now-open receptacle. The edit is accepted only after focused closed-receptacle placement cases improve and protected open-receptacle, non-openable-receptacle, and object-retrieval cases remain stable; the KB version is then advanced with provenance recorded for later inspection and rollback.

Kintsugi evolves the KB through accepted typed edits proposed from rollout evidence. The system has three training-time roles. The \emph{editor} is a restricted LLM agent that runs only between rollouts; it reads trajectory evidence, localizes the likely failure layer, and proposes a typed policy edit. The \emph{applier} deterministically checks whether the edit is schema-valid and well typed. The \emph{verifier} decides whether the resulting KB version improves focused success, or leaves focused success unchanged while improving a pre-declared trajectory-health metric, without reducing protected behavior. The editor is therefore a proposal module, not an oracle.

Let \(K_t\) be the executable KB after iteration \(t\), \(\tau_t\) the current trajectory evidence, \(C\) the source contracts, and \(V(K_t)\) the verifier summary. The editor proposes \(\Delta_t=A(\tau_t,C,V(K_t))\). Let \(M_{\mathrm{focus}}\) denote the focused endpoint metric, \(M_{\mathrm{protect}}\) the protected-regression metric, and \(H_{\mathrm{focus}}\) a pre-declared trajectory-health metric, such as fewer invalid actions, shorter successful trajectories, fewer recovery steps, or improved subgoal progress. The policy evolves only if deterministic application and validation accept the edit:
\[
K_{t+1} =
\begin{cases}
K_t \oplus \Delta_t,
&
\begin{aligned}[t]
&\text{if } \Delta_t \text{ type-checks, }
M_{\mathrm{protect}}(K_t \oplus \Delta_t) \geq M_{\mathrm{protect}}(K_t),\\
&\text{and either }
M_{\mathrm{focus}}(K_t \oplus \Delta_t) > M_{\mathrm{focus}}(K_t)\\
&\text{or }
M_{\mathrm{focus}}(K_t \oplus \Delta_t) = M_{\mathrm{focus}}(K_t)
\text{ and }
H_{\mathrm{focus}}(K_t \oplus \Delta_t) > H_{\mathrm{focus}}(K_t),
\end{aligned}
\\
K_t, & \text{otherwise.}
\end{cases}
\]
Rejected hypotheses are retained in the audit log, so failed proposals remain evidence without becoming policy, preserving a traceable record of failed alternatives.

Figure~\ref{fig:workflow} illustrates how this update rule is instantiated in the closed loop. Each iteration begins with a constrained agent running inside a sandbox whose tools and paths are scoped at the runtime level. The agent invokes a verifier or rollout runner and reads a trajectory-bank summary. Successful, failed, and partial-progress episodes are all evidence, because a success can contain waste or lucky recovery and a failure can contain a reusable subskill. Reading this evidence together with the source contracts of the active environment, the agent performs failure-layer localization: it commits to a hypothesis about which layer in Appendix Table~\ref{tab:app-kb-layers} owns the observed failure or trajectory-health issue. The hypothesis constrains the next move to a typed edit drawn from the operation vocabulary admitted by that layer, rather than a free-form policy change.

The update is then deterministically applied and schema-checked. Finally, the verifier re-runs focused validation and protected-regression checks. An update is accepted when focused success improves, or when focused success is unchanged and a pre-declared trajectory-health metric improves, with no protected regression. Otherwise the hypothesis is rejected or the KB is restored. This gating is what turns a restricted LLM agent proposal into auditable policy evolution. Appendices~\ref{app:agentic-harness} and~\ref{app:verification-details} detail the restricted harness and the smoke, focused-validation, and protected-regression gates; Appendix~\ref{app:method-details} and Table~\ref{tab:app-grammar} give the KB schema and typed mutation surface.
\section{Deployment as Zero-LLM Symbolic Execution}
\label{sec:deployment}

After training, the accepted KB is frozen and used as the deployed policy artifact. Let \(K^\star\) denote the final accepted KB. At inference step \(t\), the executor receives the object-centric state \(s_t\), the goal \(g\), and deterministically maps them to an action or skill call, \(a_t=E(K^\star,s_t,g)\). The executor \(E\) grounds KB predicates against the current state, selects applicable operators or policy schemas, checks monitors and recovery conditions, and emits the next bound skill or environment action. No LLM call is made during this mapping.

This deployment mode separates training-time diagnosis from test-time execution. During training, the restricted LLM agent proposes edit hypotheses, but at inference the actor is the deterministic executor over \(K^\star\). Thus the deployed policy is \(\pi_{K^\star}(s_t,g):=E(K^\star,s_t,g)\), rather than a language-model policy \(\pi_{\mathrm{LLM}}(s_t,g)\). The verifier controls which edits enter \(K^\star\), and the executor consumes only this accepted artifact during every deployment decision.

The executor also provides decision provenance. For each action, the system records a trace \(\mathrm{Trace}(a_t)=\{e\in K^\star: e \text{ was used to ground, select, monitor, recover, or bind } a_t\}\). Equivalently, \(\mathrm{Trace}(a_t)\subseteq K^\star\), so each deployed action can be attributed to identifiable predicates, operators, schemas, monitors, recovery rules, or skill bindings. This makes zero-LLM execution more than an efficiency choice: it is the mechanism by which the accepted KB becomes the deployed policy.

%% file: sections/04_experiments.tex
\begin{table}[t]
\caption{\textbf{Frozen-KB endpoint performance under zero-LLM execution.} Kintsugi executes the accepted KB at test time and matches or exceeds long-horizon language-agent baselines on ALFWorld, WebShop, and TextCraft. Kintsugi and access-control rows report mean \(\pm\) std over 5 seeded runs.}
\label{tab:rq1-textagent}
\centering
\setlength{\tabcolsep}{7pt}
\renewcommand{\arraystretch}{1.05}
\begin{tabular*}{\linewidth}{@{\extracolsep{\fill}}lcccc@{}}
\toprule
\textbf{Method} &
\makecell{\textbf{ALFWorld}\\\textbf{SR (\%)}} &
\makecell{\textbf{WebShop}\\\textbf{SR (\%)}} &
\makecell{\textbf{WebShop}\\\textbf{Score}} &
\makecell{\textbf{TextCraft}\\\textbf{SR (\%)}} \\
\midrule
ReAct~\citep{yao2023react} & 76.87 & 32 & 0.5010 & 62\\
Reflexion~\citep{shinn2023reflexion} & 82.66 & 35 & 0.5204 & 69\\
ADaPT~\citep{prasad2024adapt} & 72.39 & 32 & 0.5355 & 77\\
StateAct~\citep{rozanov2025stateact} & 63.43 & 17 & 0.2973 & 68\\
ExpeL~\citep{zhao2024expel} & 85.07 & 29 & 0.4582 & 88\\
WALL-E 2.0~\citep{zhou2025wall} & 82.84 & 34 & 0.5998 & 66\\
AWM~\citep{wang2024agent} & 88.81 & 32 & 0.5160 & 66\\
Dual Memory~\citep{wen2026aligning} & 94.78 & 51 & 0.7132 & 94\\
\midrule
LLM w/o KB & \(39.5{\pm}1.80\) & \(21{\pm}2\) & \(0.5819{\pm}0.0180\) & \(53{\pm}2\)\\
LLM w/ KB prompt & \(50{\pm}2.10\) & \(47{\pm}2\) & \(0.7300{\pm}0.0200\) & \(55{\pm}2\)\\
LLM w/ KB tools & \(97.80{\pm}0.90\) & \(36{\pm}1\) & \(0.7034{\pm}0.0100\) & \(97{\pm}1\)\\
\textbf{Kintsugi (ours)} & \(\mathbf{100.00{\pm}0.00}\) & \(\mathbf{52{\pm}1}\) & \(\mathbf{0.7392{\pm}0.0080}\) & \(\mathbf{100{\pm}0}\)\\
\bottomrule
\end{tabular*}
\end{table}

\section{Experiments}
We organize our experiments around five questions that build from endpoint performance to artifact evidence, local editability, cross-domain extension, and layer necessity:
\textbf{(RQ1)} Does the final KB achieve strong performance on long-horizon text-agent tasks?
\textbf{(RQ2)} Does capability emerge through verifier-gated KB evolution and reside in the executable artifact?
\textbf{(RQ3)} Can policy behavior be changed through localized typed KB edits without off-target regression?
\textbf{(RQ4)} Does the same KB-driven symbolic-policy interface extend to object-centric manipulation under compatible state and skill interfaces?
\textbf{(RQ5)} Which KB layers and executor contracts are necessary for execution, planning, recovery, and maintenance?

\paragraph{Benchmark protocols.}
For text-agent evaluation, we follow the Dual Memory-aligned~\citep{wen2026aligning} protocol used for Table~\ref{tab:rq1-textagent}: ALFWorld~\cite{shridhar2020alfworld} uses 134 valid-unseen games with a 50-step horizon, WebShop~\cite{yao2022webshop} uses 100 public-text shopping sessions for baseline comparison, and TextCraft~\cite{prasad2024adapt} uses the 100 AgentGym AgentEval item IDs with a 40-action budget. Appendix~\ref{app:benchmark-protocols} gives the exact banks, horizons, success definitions, and 500-session WebShop diagnostics.

For object-centric manipulation, MetaWorld~\citep{yu2020meta} is evaluated on the MT10 and MT50 multi-task task sets with 50 rollouts per task, a 500-step cap, and the environment binary success predicate; ``held-out'' refers to held-out reset seeds within the same MT task families, not held-out task families from the ML10/ML45 meta-learning protocol. Predicators-style bilevel-planning environments~\citep{silver2021symbolicoperators,chitnis2022learning,silver2023predicateinvention} use 10 seeds and 50 held-out evaluation tasks per environment, with success defined by symbolic goal satisfaction under the planner/executor interface. RoboSuite~\citep{zhu2020robosuite} is evaluated with simulator-state observations, fixed robot/controller settings, task-specific success predicates, a 500-step cap, and 5 reset seeds per compatible single-arm task. Appendix~\ref{app:benchmark-protocols} gives the exact object-centric protocol, and Appendix~\ref{app:rq4-details} gives the boundary evidence and failure-localization analysis.

\subsection{The Executable KB Outperforms Text-Agent Baselines}
\label{sec:textagent-baselines}

To address RQ1, we evaluate the final accepted KB on ALFWorld, WebShop, and TextCraft. These benchmarks stress complementary forms of task knowledge: object-centric household planning, reward-oriented decision making, and compositional dependency resolution. We compare against representative long-horizon language-agent baselines, including prompt-based acting and reflection methods, decomposition or state-tracking agents, experiential or workflow-memory methods, and recent neurosymbolic or world-model-style agents. These baselines are evaluated as endpoint task-solving systems; in contrast, Kintsugi uses the LLM only between rollouts to propose typed KB edits, and executes the accepted KB with zero LLM calls at test time.

Table~\ref{tab:rq1-textagent} shows that Kintsugi reaches ceiling performance on ALFWorld and TextCraft and gives a smaller WebShop gain under the fixed-artifact deployment contract. The ALFWorld result indicates that the learned KB can encode the preconditions, effects, schemas, and recovery behavior needed for long-horizon household planning. The WebShop and TextCraft results suggest that the same executable-KB formulation also supports reward-oriented selection and compositional dependency reasoning. Thus, RQ1 establishes the final task capability of the accepted KB; RQ2 later isolates how this capability is acquired and whether it resides in the executable artifact. Appendix~\ref{app:rq1-details} analyzes the baseline families and the fixed-artifact endpoint behavior behind these rows.

\subsection{Capability Resides in the Executable KB Artifact}
\label{sec:rq2-artifact}

RQ1 establishes that the final accepted KB is a strong deployed policy. To answer RQ2, we ask whether this capability is acquired through verifier-gated KB evolution and whether the information needed for action selection resides in the executable artifact. Starting from an empty task-policy scaffold, Kintsugi admits policy knowledge only through verifier-gated typed edits; on ALFWorld, three editing rounds add 13 \textsc{ADD\_RULE} edits and improve the artifact from \(0.0\%\) to \(100.0\% \) success. This cold-start trajectory shows that task competence is accumulated through accepted KB changes over successive verifier-gated rounds rather than inherited from a pretrained test-time actor.

We then freeze the completed KBs and vary only how an actor accesses them, holding the underlying policy knowledge fixed. As shown in Table~\ref{tab:rq1-textagent}, both KB-access settings improve over a no-KB language actor on most metrics, and typed KB-tool access is especially strong on ALFWorld and TextCraft. These rows show that the accepted KB contains substantial task knowledge that can guide an external language actor. However, direct KB execution performs best overall: Kintsugi runs the accepted KB itself as the policy, rather than asking an LLM to interpret it as context or retrieved facts. This gap supports the artifact-centered claim. The KB is not only informative, but executable: the task knowledge needed for action selection is encoded in a form that the symbolic executor uses more reliably than a test-time language model. Thus, capability is acquired through verifier-gated KB edits, becomes accessible as useful task knowledge to an LLM, and is deployed most effectively when the accepted KB is executed directly with zero LLM calls. Appendix~\ref{app:rq2-details} expands this result into the cold-start ledger, access-control analysis, and task-type diagnostics.
\subsection{Localized KB Edits Repair Target Behaviors}
\label{sec:localized-kb-edits}

To address RQ3, we test whether policy behavior can be changed through localized typed KB updates rather than prompt rewriting, free-form code patches, or parameter updates. The ALFWorld cold-start ledger provides a clean editability test: each failure trace exposes named rules, predicates, states, and admissible actions, making it possible to localize a missing transition, apply a small KB edit, and verify whether the target behavior improves without disrupting behavior that was already solved.

Table~\ref{tab:rq3-editability} shows that the accepted edits are local in their effect. The first edit round repairs the basic pick/place transition family; the second repairs the light-use transition; the third repairs the tool-processing chains for clean, heat, and cool tasks. In each case, the target task family moves from complete failure to complete success, while previously solved families show no off-target regression.

\begin{wraptable}{r}{0.58\textwidth}
\vspace{-8pt}
\centering
\caption{\textbf{Local KB edits repair behavior without protected regression.}}
\label{tab:rq3-editability}
\setlength{\tabcolsep}{3pt}
\renewcommand{\arraystretch}{1.06}
\begin{tabular*}{0.58\textwidth}{@{\extracolsep{\fill}}lccc@{}}
\toprule
\textbf{Edit} & \textbf{Target repair} & \textbf{Target} & \textbf{Reg.} \\
\midrule
0$\rightarrow$1 & pick/place chain & \(0/41 \rightarrow 41/41\) & \(0.0\) pp \\
1$\rightarrow$2 & light-use transition & \(0/18 \rightarrow 18/18\) & \(0.0\) pp \\
2$\rightarrow$3 & clean/heat/cool chain & \(0/75 \rightarrow 75/75\) & \(0.0\) pp \\
\bottomrule
\end{tabular*}
\vspace{-8pt}
\end{wraptable}

This supports the white-box maintenance claim at the policy-knowledge level. The policy can be inspected through named rules and predicates, the missing transition can be described as a localized KB repair, and the verifier accepts the update only when the focused target improves without breaking protected behavior. Thus behavior changes through small, auditable KB edits rather than opaque retraining or test-time prompt adjustment. Appendix~\ref{app:rq3-details} gives the edit loci, a concrete repair example, and the accepted-edit ledger behind this summary.

\subsection{KB-Driven Policies Extend to Object-Centric Manipulation}
\label{sec:rq4-object-centric}

\begin{figure}[t]
\centering
\includegraphics[width=0.98\linewidth]{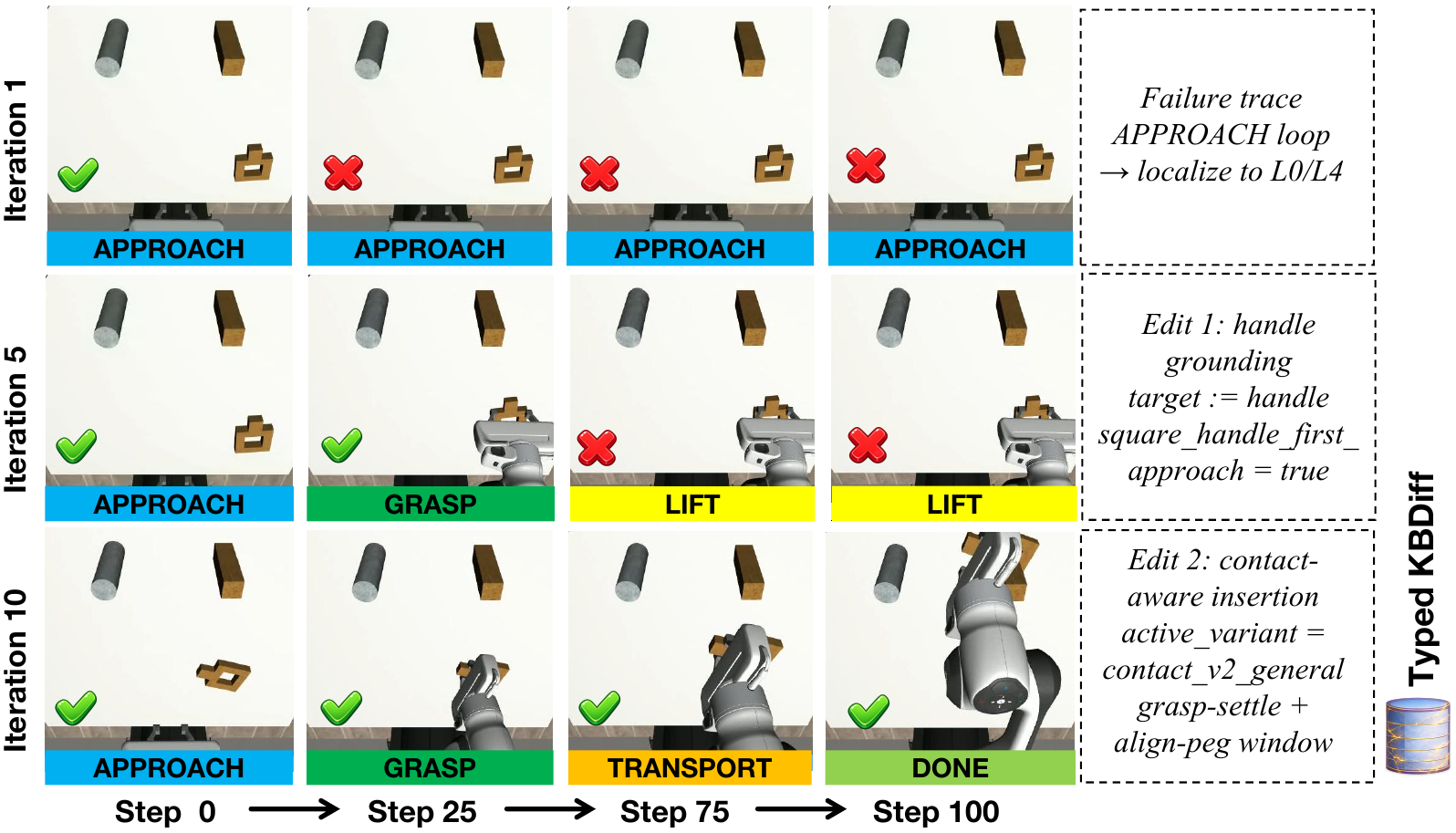}
\caption{\textbf{Kintsugi converts failure traces into typed KB edits.} In SquareNut, each iteration localizes the current rollout failure to editable KB entries: early APPROACH-loop traces produce handle-first grounding, while later contact/insertion failures produce a contact-aware insertion variant. The accumulated edits let the zero-LLM executor progress through GRASP, TRANSPORT, and DONE, showing iterative trace-to-edit policy evolution.}
\label{fig:rq4-robosuite-squarenut}
\end{figure}

To answer RQ4, we test whether the same KB-driven symbolic policy interface extends beyond text agents to object-centric manipulation. Kintsugi does not learn low-level motor control from scratch; instead, it maps object-centric state to manipulation behavior through grounded predicates, operators, schemas, monitors, recoveries, and skill/action bindings.

We evaluate MetaWorld, Predicators, and RoboSuite. MetaWorld tests broad single-arm manipulation and held-out reset robustness; Predicators tests symbolic-induction-style manipulation outside text; and RoboSuite tests robot-related manipulation under true policy-action rollouts. MetaWorld uses \(n{=}50\) rollouts per task, Predicators uses \(10\) seeds \(\times\) \(50\) tasks per environment, and RoboSuite uses 15 compatible single-arm tasks with \(n{=}5\) seeds per task for the reported aggregate result. Success is reported as mean \(\pm\) standard deviation over the declared endpoint success-rate units. RoboSuite is a boundary study because it tests the symbolic-to-continuous skill interface under simulator-state single-arm rollouts, not an official benchmark ranking.

\begin{wraptable}{r}{0.56\textwidth}
\vspace{-10pt}
\centering
\caption{\textbf{Kintsugi extends to object-centric manipulation tasks.}
RoboSuite is reported as transitional boundary evidence, not as a canonical
official RoboSuite benchmark comparison.}
\label{tab:rq4-object-centric}
\setlength{\tabcolsep}{3pt}
\renewcommand{\arraystretch}{1.06}
\begin{tabular*}{0.56\textwidth}{@{\extracolsep{\fill}}lcc@{}}
\toprule
\textbf{Suite} & \textbf{Solved} & \textbf{Success} \\
\midrule
MT-10 held-out & \(496/500\) & \(99.2{\pm}1.93\%\) \\
MT-50 held-out & \(2458/2500\) & \(98.32{\pm}4.13\%\) \\
Predicators & \(1997/2000\) & \(99.85{\pm}0.34\%\) \\
RoboSuite single-arm & \(68/75\) & \(90.7{\pm}1.37\%\) \\
\bottomrule
\end{tabular*}
\vspace{-10pt}
\end{wraptable}

Table~\ref{tab:rq4-object-centric} shows that the KB-driven interface extends beyond language-only environments. MetaWorld and Predicators indicate that typed predicates, operators, and schemas support broad object-centric manipulation. RoboSuite provides a stricter robot-related check under true policy-action rollouts. The remaining failures concentrate in harder cases, most notably \texttt{tool\_hang}. Appendix Table~\ref{tab:app-robosuite-aggregate} gives the per-task breakdown, and Appendix~\ref{app:rq4-details} gives evidence that these drops arise at the symbolic-to-continuous execution boundary: grounding, recovery, scene-conditioned sequencing, and skill/action binding. Figure~\ref{fig:rq4-robosuite-squarenut} gives a qualitative RoboSuite example. In SquareNut, Kintsugi converts successive rollout failures into localized KB edits rather than a single monolithic repair. Early traces stall in \texttt{APPROACH}, exposing a handle-grounding failure and yielding a handle-first approach edit. Later traces reveal contact and insertion failures, localized to the policy-schema layer and repaired by a contact-aware insertion variant. The accumulated edits let the same zero-LLM executor progress through \texttt{GRASP}, \texttt{TRANSPORT}, and \texttt{DONE}. This illustrates the RQ4 boundary: Kintsugi can drive object-centric manipulation when state and skill interfaces are compatible, while RoboSuite remains a KB-guided symbolic-control protocol rather than a full visual or real-robot learning claim.
\subsection{Typed KB Layers Are Load-Bearing Contracts}
\label{sec:rq5-ablation}

To answer RQ5, we freeze the final accepted ALFWorld KB and intervene on individual \(S0/L0\)--\(L7\) layers and executor contracts, without retraining or LLM calls. The modified artifact is then executed by the same deterministic executor. Detailed results are reported in Appendix~\ref{app:rq5-details} and Appendix Tables~\ref{tab:appendix-layer-ablation} and~\ref{tab:component-ablation}; here we summarize the main failure signatures. The interventions produce structured rather than uniform degradation. Removing the source--grounding--predicate interface disables executable state access; removing ontology and affordances breaks action applicability; removing policy schemas disables action selection; and removing goals erases task decomposition. Operator, monitor, recovery, and experience ablations are less visible in clean rollouts but expose targeted failures in causal-effect reasoning, stress handling, auditability, and rollback. These patterns show that the typed layers are not decorative labels: they define separable contracts for state access, grounding, predicates, world semantics, causal effects, action selection, recovery, provenance, and goals, which is what makes failures localizable and repairs verifier-gated.

%% file: sections/02_related_work.tex
\section{Related Work}

\paragraph{Symbolic induction from interaction.}
The closest comparator is work on learning symbolic structure for task and motion planning. ~\cite{konidaris2018skillstosymbols} ground low-level skills into symbolic predicates, and ~\cite{garrett2021tamp} review how planners combine discrete search with continuous samplers. ~\cite{silver2021symbolicoperators} learn symbolic operators for TAMP, while ~\cite{silver2023predicateinvention} invent predicates from demonstrations. Related lifelong, language-guided, and visual predicate systems extend this direction through experience, language feedback, visual predicates, and exogenous abstract models~\cite{mendez2023embodiedlifelong,han2024interpret,li2025visualpredicator,liang2026exopredicator}. Our difference is the unit and update path of learning: Kintsugi treats the full executable policy stack as a typed KB, including grounding, affordances, operators, schemas, monitors, recovery rules, experience, and goals. Each accepted update is a localized typed edit induced from traces and accepted by validation, rather than a learned predicate set, operator learner, or domain-specific abstraction search.
\paragraph{LLM-as-policy and LLM-augmented symbolic methods.}
Many agents place the LLM inside the inference loop: ReAct~\cite{yao2023react} introduced thought-and-action prompting, Reflexion~\cite{shinn2023reflexion} added verbal reinforcement, Voyager~\cite{wang2024voyager} used an LLM-authored skill library, and AutoManual~\cite{chen2024automanual} induced actionable manuals through interaction. Other systems externalize reasoning into memory~\cite{packer2023memgpt,sumers2023cognitive} or synthesize executable code and symbolic plans~\cite{liang2023codeaspolicies}. These approaches improve inspectability or deterministic inference, but their artifacts are often prompt memories, memory extensions, free-form programs, or one-shot symbolic files. Kintsugi instead uses the LLM only between rollouts to propose structured edits; the typed KB, applier, executor, and verifier decide whether an edit becomes policy.
\paragraph{VLA and world-model policies.}
Vision-Language-Action models such as RT-2~\cite{zitkovich2023rt}, OpenVLA~\cite{kim2024openvla}, Octo~\cite{team2024octo}, $\pi_0$~\cite{black2024pi_0}, and $\pi_{0.7}$~\cite{physicalintelligence2026pi07} excel at data-scale perception and open-vocabulary control. World-model methods such as Dreamer~\cite{hafner2023mastering} and TD-MPC2~\cite{hansen2024tdmpc2} learn latent dynamics, while diffusion policies such as DP3~\cite{ze2024dp3} generate actions over 3D observations. Kintsugi is complementary rather than directly comparable: neural perception and motor policies can serve as perception sources or skill bindings, while Kintsugi edits and verifies the symbolic task knowledge above them.

%% file: sections/06_conclusion.tex
\section{Conclusion}

Kintsugi reframes embodied policy learning as white-box policy evolution. Instead of storing improvements in opaque parameters, prompt memories, or unstructured patches, Kintsugi maintains an executable typed KB whose decisions are traceable and whose updates are localized, validated, and replayable. The LLM only proposes typed edits between rollouts; deterministic application and verifier evidence decide whether policy changes. At inference, the accepted KB executes symbolically with no LLM in the loop. For object-centric tasks with calibrated skill interfaces, failures become localized edit hypotheses, and accepted edits become persistent policy knowledge. Kintsugi does not replace neural perception or motor policies, but provides a complementary white-box layer for maintaining, debugging, and reusing task knowledge above them.

%% file: sections/A_appendix.tex

\section{Appendix Roadmap and Evaluation Contract}
\label{app:protocol}

The appendix supplements the main text rather than introducing a separate
story. Section~\ref{app:method-details} expands the policy-artifact and
reasoning-site definitions used by the method.
Section~\ref{app:agentic-harness} describes the restricted agentic training
harness, and Section~\ref{app:verification-details} explains the smoke,
focused-validation, and protected-regression gates.
Section~\ref{app:formal-properties} states the limited formal guarantees that
follow from typed application and verifier-gated acceptance.
Section~\ref{app:benchmark-protocols} records the benchmark banks, action
surfaces, horizons, and success metrics used by the paper-facing results.
Sections~\ref{app:rq1-details}--\ref{app:rq5-details} then mirror the five
experiment subsections: executable KB endpoint performance, capability in the
executable KB artifact, localized KB edits, object-centric manipulation, and
load-bearing typed layers.

\paragraph{Shared evaluation contract.}
All Kintsugi evaluation rows use a fixed-artifact deployment contract. The policy
artifact is frozen before test evaluation; the actor is the accepted KB plus a
deterministic executor; no LLM call is made during test-time action selection;
and success is measured by the benchmark or environment evaluator rather than
by symbolic self-certification. Candidate knowledge enters the policy artifact
only through typed edits accepted by deterministic application and verifier
gates. Debug runs, diagnostic upper bounds, and transitional adapter studies
are kept separate from main comparable claims.

\paragraph{Environment-interface scope.}
Kintsugi does not assume that an editor must rediscover an environment API from
raw bytes. Each benchmark supplies a public interface contract: observation
adapter, action or skill binding, success verifier, and the typed mutation
surface that a KB edit may target. This is analogous to giving a robot its
manual: the policy learner must know what sensors mean, which actions or skills
can be called, what arguments they accept, and how success is measured before it
can learn a reliable task policy above that interface. In text environments this
contract may also include task-family or policy-schema requirements, such as the
public ALFWorld action vocabulary and the high-level schemas that the
deterministic executor can bind to admissible text actions. These requirements
are not episode answers: they do not reveal object locations, hidden product
labels, or successful trajectories. The learned object remains the accepted
KB/policy artifact authored through KBDiff and validated by rollout evidence.
Stronger editor models may infer useful schemas from weaker contracts, but every
deployed capability must still enter the same typed artifact and pass the same
verifier gate. Thus Kintsugi is general at the controller and mutation level,
while a new benchmark still needs an adapter that exposes public state, actions
or skills, and the benchmark success signal. The adapter is substrate, not
learned policy content.
\begin{quote}\small
\begin{verbatim}
run_loop(K, benchmark_adapter):
  pre    = benchmark_adapter.evaluate(K)
  diff   = propose_kbdiff(summarize(pre.traces), localize(pre), K)
  K'     = apply_diff(K, diff)        # deterministic schema/type gate
  post   = benchmark_adapter.evaluate(K')
  keep K' iff protected does not regress and
             (focused success improves or
              focused success is unchanged and
              pre-declared trajectory health improves)
\end{verbatim}
\end{quote}

\paragraph{What the appendix adds.}
The main text states the white-box claim compactly. Here we unpack the
observable requirements behind that claim. The policy state must be inspectable
as typed KB entries rather than only as parameters or prompt text. A deployed
action must have a trace to the predicates, operators, policy schemas,
monitors, recovery rules, or skill bindings that produced it. A learned update
must name a layer, key, operation, payload, evidence source, and validation
metric. Finally, a candidate update must remain a hypothesis until it passes
type checking, focused validation, and protected-regression checks. The
following sections show these requirements in the concrete ALFWorld artifact
and then report the detailed protocols and results behind each RQ.

\FloatBarrier

\section{Method Details: Policy Artifacts and Reasoning Sites}
\label{app:method-details}

This section expands the method definitions from the main text. Kintsugi's policy
state is an executable artifact \(K\), not an unstructured prompt memory. The
artifact names an observation substrate \(S0\) and typed policy-knowledge
layers \(L0\)--\(L7\). The layer vocabulary is used in three places: to execute
the policy, to localize failures, and to constrain admissible updates.
Table~\ref{tab:app-kb-layers} summarizes the editable hierarchy.

\begin{table}[t]
\caption{\textbf{Kintsugi's editable policy knowledge hierarchy.} \(S0\) defines
the observation substrate, while \(L0\)--\(L7\) provide typed edit loci for
grounding, predicates, operators, policies, recovery, experience, and goals,
making failures localizable and repairs auditable.}
\label{tab:app-kb-layers}
\centering
\small
\setlength{\tabcolsep}{3pt}
\renewcommand{\arraystretch}{1.08}
\begin{tabularx}{\linewidth}{@{}L{0.08\linewidth}L{0.20\linewidth}Y@{}}
\toprule
\textbf{Layer} & \textbf{Role} & \textbf{Representative contents or updates}\\
\midrule
\(S0\) & Source &
Oracle or simulator state, tracker output, robot estimator, or visual
perception module; declares exposed state information.\\
\(L0\) & Grounding &
Object schemas, calibration, derived geometry, and target functions; update
grounding rules or target functions.\\
\(L1\) & Predicates &
Grounded facts over objects, relations, contacts, geometry, and task states;
add predicates or adjust thresholds.\\
\(L2\) & Ontology &
Object types, affordances, spatial priors, and articulation types; add object
facts or update affordances.\\
\(L3\) & Operators &
Symbolic preconditions, effects, parameters, and skill bindings; add operators
or modify preconditions/effects.\\
\(L4\) & Policy schemas &
State machines, options, subgoal orderings, and skill-selection logic; add
schemas or rebind skills.\\
\(L5\) & Monitors / recovery &
Success checks, failure detectors, timeout conditions, and recovery policies;
add monitors or recovery rules.\\
\(L6\) & Experience &
Accepted/rejected hypotheses, trajectory evidence, provenance records, and
verifier outcomes; append episodes or audit logs.\\
\(L7\) & Goals &
Task decompositions, goal templates, terminal conditions, and success
specifications; add task schemas or update goals.\\
\bottomrule
\end{tabularx}
\end{table}

\paragraph{Layer contents in the artifact.}
The layer names are not only conceptual labels; they correspond to concrete
fields read by the implementation. \(S0\) names the observation source and the
contract by which state enters the executor. In ALFWorld this is the TextWorld
state/admissible-command interface; in manipulation settings it may be
simulator state, a tracker, or a calibrated robot estimator. \(L0\) grounding
turns that source into typed object and relation fields, such as current
receptacle, inventory, known receptacles, visible contents, and open/closed
state. \(L1\) predicates expose Boolean tests over those grounded fields:
\texttt{holds\_target}, \texttt{at\_goal\_recep},
\texttt{current\_openable\_closed}, \texttt{at\_process\_tool}, and
\texttt{can\_process\_here} are ALFWorld examples.

\(L2\) stores world semantics: receptacle openability, process tools,
affordances, object categories, confusable classes, and spatial priors. \(L3\)
stores symbolic operator effects, such as the pre/post conditions for
\texttt{take}, \texttt{open}, \texttt{put}, \texttt{clean}, \texttt{heat}, and
\texttt{cool}. \(L4\) stores executable policy schemas: priority rules,
state-machine phases, or option-selection logic. \(L5\) stores or exposes
monitor and recovery contracts, such as success checks, failure detectors, and
fallback behavior. In ALFWorld, part of this recovery contract is implemented
by the executor's admissible-command fallback rather than learned as a new task
rule. \(L6\) stores provenance: accepted and rejected hypotheses, rollout
traces, verifier outcomes, and audit records. \(L7\) stores goal templates and
task decompositions, such as the sequence for
\texttt{pick\_heat\_then\_place\_in\_recep}.

\begin{quote}\small
\textbf{ALFWorld layer example.}
\(S0\): TextWorld observation and admissible actions.
\(L0\): current receptacle, inventory, known objects, open-state map.
\(L1\): \texttt{ready\_to\_deposit}, \texttt{at\_goal\_recep},
\texttt{current\_openable\_closed}.
\(L2\): \texttt{microwave.tool\_for=heat},
\texttt{fridge.tool\_for=cool}, \texttt{cabinet.openable=true}.
\(L3\): \texttt{put} requires holding the object and the target receptacle
being open if openable.
\(L4\): \texttt{OpenGoalRecep} and \texttt{DepositHeld} rules.
\(L5\): executor-side admissible-command fallback and recovery contract.
\(L6\): accepted-edit ledger and validation result.
\(L7\): heat-then-place task schema.
\end{quote}

\paragraph{Reasoning sites.}
Kintsugi deliberately separates three kinds of reasoning that are often conflated
in language-agent systems. The restricted LLM editor performs diagnostic
reasoning between rollouts: it reads compact trajectory evidence and source
contracts, then proposes a failure layer and typed edit hypothesis. This output
is not trusted as policy. The deterministic executor performs policy reasoning
at deployment: given the accepted KB and current state, it evaluates predicates,
fires rules or schemas, checks monitors, and emits a skill or environment
action. The verifier performs acceptance reasoning: it decides whether the
candidate artifact satisfies focused acceptance---either through improved
focused success or through unchanged focused success with improved
pre-declared trajectory health---without violating the protected set. Thus the
LLM can help generate hypotheses, but it neither acts at test time nor decides
which hypotheses become policy.

\paragraph{KBDiff as the mutation channel.}
At the paper level, a candidate update is the typed edit
\[
\Delta = (\mathrm{layer}, \mathrm{key}, \mathrm{op}, \mathrm{payload},
\mathrm{evidence}, \mathrm{metric}, \mathrm{regression\_set}).
\]
The implementation represents the mutation-facing part of this edit as a typed
\texttt{KBDiff} record.
The \texttt{op} field is drawn from a closed vocabulary such as
\texttt{add\_predicate}, \texttt{modify\_threshold}, \texttt{add\_rule},
\texttt{modify\_rule\_guard}, \texttt{add\_skill},
\texttt{add\_object\_fact}, \texttt{add\_task\_schema},
\texttt{add\_operator\_schema}, \texttt{add\_policy\_schema}, and
\texttt{add\_monitor}. The \texttt{path} is a dot-separated location inside the
YAML artifact, such as \texttt{procedural.rules} or
\texttt{semantic.spatial\_priors.cup}. The \texttt{payload} must match the
operation-specific schema. The deterministic applier loads the YAML in
round-trip mode, checks the path and payload contract, mutates the in-memory KB,
and returns both the new YAML text and an audit record containing the operation,
path, rationale, expected effect, and before/after snippets. If the path is
missing or the payload is malformed, application fails before any environment
validation.

\begin{quote}\small
\textbf{Example KBDiff for the closed-receptacle repair.}\\
\texttt{op: add\_rule}\\
\texttt{path: procedural.rules}\\
\texttt{payload.rule.name: OpenGoalRecep}\\
\texttt{payload.rule.cond: all(task\_uses\_deposit, ready\_to\_deposit,
at\_goal\_recep, current\_openable\_closed)}\\
\texttt{payload.rule.action: OPEN}\\
\texttt{rationale: At the goal receptacle with the held target, closed
openable containers must be opened before deposit.}\\
\texttt{expected\_effect: closed-container deposit cases improve without
changing non-openable deposit cases.}
\end{quote}

\begin{table}[t]
\caption{\textbf{Typed update grammar used by the verifier-gated applier.}}
\label{tab:app-grammar}
\centering
\small
\setlength{\tabcolsep}{3pt}
\renewcommand{\arraystretch}{1.08}
\begin{tabularx}{\linewidth}{@{}L{0.12\linewidth}L{0.22\linewidth}L{0.35\linewidth}Y@{}}
\toprule
\textbf{Layer} & \textbf{Key type} & \textbf{Representative operations} & \textbf{Required checks}\\
\midrule
\(S0\) & Source contract &
Declare source; bind observation field; set read-only source metadata &
Source contract exists and learned-policy claims do not mutate oracle state.\\
\(L0\) & Grounding key &
Add or update grounding; calibrate target function; bind visual or simulator
field to object slot &
Payload declares object type, units, source field, and uncertainty or threshold
when applicable.\\
\(L1\) & Predicate key &
Add predicate; add derived relation; change bounded threshold &
Typed arguments resolve to grounded fields; thresholds are bounded and
verifier-visible.\\
\(L2\) & Type or affordance key &
Add object fact; update affordance; add process-tool or spatial prior &
Object class is known or declared; affordance is consumed by a valid operator or
schema.\\
\(L3\) & Operator key &
Add operator; add or remove precondition; add effect; bind skill arguments &
Preconditions and effects resolve to predicates; emitted skill/action type-checks.\\
\(L4\) & Policy schema key &
Add state machine; add rule; change priority; rebind skill; update termination &
Rule guards resolve to predicates; priority is explicit; termination condition exists.\\
\(L5\) & Monitor/recovery key &
Add success check; add timeout/no-progress detector; add recovery rule &
Trigger, repair target, and protected regression scope are declared.\\
\(L6\) & Evidence key &
Append trajectory id; record accepted or rejected hypothesis; store verifier outcome &
Evidence id, focused metric, protected metric, and status are present.\\
\(L7\) & Goal key &
Add task schema; update decomposition; add terminal condition &
Decomposition resolves to operators or policy schemas; terminal condition is
verifier-measurable.\\
\bottomrule
\end{tabularx}
\end{table}

\paragraph{Applier and verifier checks.}
The grammar table is enforced before a candidate reaches environment
validation. The applier rejects edits whose operation is not allowed for the
target layer, whose path does not exist, whose payload has missing fields, or
whose rule references undefined predicates, skills, objects, or goal schemas.
For example, an \texttt{add\_rule} KBDiff must supply a rule name, priority or
insertion policy, condition, action schema, rationale, and expected effect.
The condition must be expressible over \(L1\) predicates or declared helper
tests; the action schema must resolve to a bound skill or admissible command;
and the expected effect determines the focused validation bank. Only after
these structural checks pass does the verifier run rollouts.

\paragraph{Committed KB entry example.}
After validation, an accepted edit is no longer just a diff; it becomes a KB
entry used by the executor. A simplified committed operator/policy entry for
object transfer has the following structure:
\begin{quote}\small
\textbf{Name:} \texttt{MoveObjectToReceptacle}.\\
\textbf{Layers touched:} \(L2\) openability fact, \(L3\) place operator,
\(L4\) deposit policy schema, \(L5\) closed-receptacle recovery monitor.\\
\textbf{Preconditions:} target object is held; target receptacle is reachable;
if the receptacle is openable, it is open before \texttt{PUT}.\\
\textbf{Effects:} object is at target receptacle; hand is empty; deposit
subgoal is satisfied.\\
\textbf{Bound skills/actions:} navigate, open, put.\\
\textbf{Provenance:} induced from failed closed-container deposit trajectories;
accepted after focused placement validation with protected open-container and
non-openable receptacle regressions unchanged.
\end{quote}
This is the concrete reason the KB is auditable: the same entry records what
state must be true, which action is emitted, what effect is expected, and which
evidence justified the edit.

\paragraph{From diff to symbolic policy.}
The training controller calls this path inside a closed loop:
\[
\begin{aligned}
&\textsc{Verify}(K_t)\rightarrow
\textsc{Summarize}(\tau_t)\rightarrow
\textsc{Localize}(\ell)\\
&\rightarrow
\textsc{Propose}(\Delta_t)\rightarrow
\textsc{Apply}(K_t,\Delta_t)\rightarrow
\textsc{Verify}(K_t').
\end{aligned}
\]
After a KBDiff is applied, it is not immediately trusted. The controller writes
the candidate KB, reruns focused validation and protected-regression checks,
and keeps the new artifact only if the verifier accepts it. Otherwise the
previous KB snapshot is restored and the rejected hypothesis remains only in
the audit log.

At deployment, the executor does not read the KBDiff log directly. It reads the
accepted KB. In ALFWorld, the rule engine loads \texttt{procedural.rules},
sorts rules by numeric priority, evaluates each rule's declarative condition
against \(L1\) predicates over the grounded state, and returns the first
matching action schema. The skill map then converts that schema into an
environment command, such as \texttt{go to microwave 1},
\texttt{open cabinet 1}, or \texttt{move cup 2 to cabinet 1}. The environment
step produces the next observation; grounding updates \(L0/L1\); and the loop
repeats. This is how the accepted KB becomes the policy \(a_t=E(K,s_t,g)\).

\paragraph{Concrete ALFWorld artifact.}
In ALFWorld, the KB is not a monolithic list of rules. It separates relatively
stable world semantics from executable control. The semantic portion stores
object and receptacle facts such as whether a receptacle is openable, which
tool realizes a process verb, and where object types are likely to be found.
For example, the KB records that a microwave is openable and is the tool for
\texttt{heat}, a fridge is openable and is the tool for \texttt{cool}, and a
sinkbasin is the tool for \texttt{clean}. It also stores task schemas: a
\texttt{pick\_heat\_then\_place\_in\_recep} task decomposes into finding the
target object, taking it, navigating to a microwave, opening the microwave,
heating the object, navigating to the goal receptacle, opening it if necessary,
and depositing the object.

The procedural portion of the same KB stores the deployed decision policy. It
uses a descending-priority first-match rule policy: the executor evaluates typed
conditions over the current grounded state and fires the first rule whose
condition is true. A representative rule is \texttt{OpenGoalRecep}: if the task
requires deposit, the held target is ready to deposit, the agent is at the goal
receptacle, and that receptacle is closed and openable, the action schema is
\texttt{OPEN}. The skill binding then emits the admissible text command, such
as \texttt{open cabinet 1}. This is the main distinction between a readable note
and an executable artifact: the same entry is used for inspection, action
selection, and later edit localization.

\begin{quote}\small
\textbf{Rule:} \texttt{OpenGoalRecep}.\\
\textbf{Condition:} task uses deposit; held target is ready to deposit; agent
is at the goal receptacle; current receptacle is openable and closed.\\
\textbf{Action schema:} \texttt{OPEN}.\\
\textbf{Purpose:} open a closed goal receptacle before attempting
\texttt{DepositHeld}.
\end{quote}

\paragraph{One decision trace.}
Consider a heat-then-place task whose target is a cup and whose final
receptacle is a cabinet. The observation source \(S0\) exposes the TextWorld
state and admissible actions; \(L7\) selects the heat-then-place schema; \(L2\)
identifies \texttt{microwave} as the heat tool and \texttt{cabinet} as
openable; \(L1\) predicates track whether the target is held, whether the agent
is at the microwave or cabinet, and whether the current receptacle is open. The
policy schema then follows the trace:
\begin{quote}\small
\texttt{SearchUnvisited} \(\rightarrow\) \texttt{GrabTarget}
\(\rightarrow\) \texttt{GoToProcessTool} \(\rightarrow\)
\texttt{ProcessHere} \(\rightarrow\) \texttt{TransportToGoal}
\(\rightarrow\) \texttt{OpenGoalRecep} \(\rightarrow\)
\texttt{DepositHeld}.
\end{quote}
Each arrow is a rule decision over named predicates, not a post-hoc natural
language explanation. If the rollout fails, the trace narrows the repair locus:
failure to find the cup points to grounding or spatial priors; failure to heat
points to process-tool semantics or skill binding; failure to deposit into a
closed cabinet points to the open-before-deposit rule or openability predicate.
\section{Agentic Training Harness}
\label{app:agentic-harness}

The agentic component is a training harness around the typed KB, not an
unconstrained code-writing agent. Its purpose is to make the LLM follow the
Kintsugi protocol: run the current artifact, read trajectory evidence, localize the
dominant failure to an editable layer, propose one typed KB edit, apply it
through the deterministic applier, and rerun the verifier. The harness is what
turns general LLM reasoning into a bounded policy-improvement procedure.

\paragraph{Sandboxed workspace.}
For cold-start runs, the driver creates a run directory containing an empty or
scaffold KB, a notes file, a verifier command, and a curated read-only context
mirror. The context mirror contains the shared Kintsugi contracts
(\texttt{kbdiff.py}, \texttt{applier.py}, \texttt{verifier.py},
\texttt{verify\_kb.py}, and related validators) plus environment-specific source
contracts. In ALFWorld, those environment files include grounding, predicates,
rules, runner, skills, vocabulary, and KB loading code. Production solution
KBs, historical result directories, and unrelated repository files are not
placed in scope. The tool runtime further restricts the agent: it may read
inside the run directory, write only \texttt{kb.yaml} and \texttt{notes.md}, and
run only the exact verifier command materialized by the harness.

\paragraph{What the LLM sees.}
The prompt tells the agent that the goal is live rediscovery from current
failure traces rather than replaying a historical KB. It gives the target
success rate, the allowed verifier command, the path of the active KB, the path
of the verifier output, and the location of curated source contracts. The LLM
therefore sees the same execution interface that the symbolic policy uses:
which predicates exist, which skills are declared, what command forms they
emit, how rules are evaluated, and how failures are summarized. This is
important because the LLM is asked to propose a typed edit against the real
executor contract, not to invent a separate policy in natural language.

\paragraph{LLM model disclosure.}
We distinguish the training-time editor model from test-time actors. The
paper-facing accepted artifacts in the headline text-agent rows are produced by
a stronger restricted editor, \texttt{gpt-5.5}, and then frozen before
deployment. We additionally report \texttt{gpt-4o} editor diagnostics to test
whether the same agentic loop still constructs useful artifacts with a weaker
editor. In all cases, the accepted KB produced by an editor run is executed
with zero LLM calls at deployment. The ALFWorld LLM-access control rows use
\texttt{gpt-4o} for the test-time language actor that interprets either
prompt-rendered KB content or KB memory tools. Local scripts may expose other
defaults for ad hoc smoke tests, but paper-facing runs should be read from
their run manifests. For RoboSuite boundary studies, the model identifier is
not a test-time policy parameter because the executed policy is the symbolic
executor; diagnostic or agentic runs record their model in the run manifest
when an LLM is used to propose edits.

\paragraph{Protocol guidance.}
The harness pushes the agent through a fixed sequence. First, it runs the
verifier and reads the compact JSON result plus sampled failure traces. Second,
it writes a short diagnosis in \texttt{notes.md}: what failed, which rule fired,
which predicate or skill was missing, and which layer owns the gap. Third, it
proposes one KBDiff whose operation and payload match the allowed mutation
schema. Fourth, it applies the edit and reruns the verifier. If the edit fails
to apply, the applier error is evidence for a corrected diff, not permission to
bypass the schema. If the edit runs but does not satisfy focused acceptance, or
if it causes protected regression, the candidate is rejected or revised in the
next iteration.

\paragraph{Role of skills.}
The harness exposes primitive skills as contracts rather than as free-form code
targets. In ALFWorld, skills are text-action emitters such as \texttt{GOTO},
\texttt{OPEN}, \texttt{TAKE}, \texttt{PUT}, \texttt{PROCESS\_HERE}, and
\texttt{GOTO\_PROCESS\_TOOL}; they translate symbolic action schemas into
admissible environment commands. In manipulation settings, skills or robot
bindings expose calibrated primitives and action-channel assumptions. The LLM
can propose a KB edit that selects, reorders, guards, or binds these skills, but
an accepted knowledge update must be distilled into the typed artifact. A
task-local Python patch can be useful as a prototype or adapter repair, but it
is not counted as learned policy knowledge until the behavior is represented as
a typed KB entry and passes validation.

\paragraph{Harness outputs.}
Each run records the prompt, context manifest, active KB, verifier outputs,
sampled traces, notes, accepted/rejected hypotheses, and final summary. The
driver also exports a paper-readable KB snapshot that reorganizes the
executable artifact into the \(S0/L0\)--\(L7\) presentation used in this paper.
These outputs are useful for audit: a reviewer can see which information the
agent had, what edit it proposed, whether the deterministic applier accepted
the edit, which verifier window was used, and why the candidate was kept or
rejected.

\paragraph{Why the harness matters.}
Without the harness, an LLM could solve a task by writing arbitrary glue code,
remembering a prompt instruction, or overfitting a seed-specific trace. With the
harness, the LLM's degrees of freedom are intentionally narrowed. It may reason
about failures and propose hypotheses, but the policy changes only through
typed KBDiff updates and verifier evidence. This is what makes the training
loop agentic without making the deployed policy an LLM policy.

\FloatBarrier

\section{Verification Details: Smoke, Focused Validation, and Regression}
\label{app:verification-details}

Kintsugi uses verification at several granularities. These checks serve different
purposes and should not be conflated. A smoke test is an inexpensive executable
sanity check; it is not a paper result and it is not sufficient for accepting a
policy update. A focused validation set tests whether the targeted behavior
actually improves. A protected regression set tests whether previously solved
behavior remains stable. Only the latter two decide whether candidate knowledge
enters the accepted KB.

\paragraph{Applier smoke tests.}
The first gate is deterministic and does not touch an environment. Given a
\texttt{KBDiff}, the applier checks that the operation is in the closed
vocabulary, the dotted path exists or is valid for the operation, and the
payload satisfies the operation-specific schema. The offline smoke tests cover
representative mutation primitives such as \texttt{modify\_threshold},
\texttt{add\_predicate}, \texttt{add\_rule}, \texttt{modify\_rule\_guard},
\texttt{extend\_skill\_body}, \texttt{add\_object\_fact}, and
\texttt{modify\_priority}. They also test that malformed payloads raise a
schema error and that the YAML round trip preserves a structurally equivalent
artifact. The benefit is immediate: many LLM-generated edit mistakes are caught
before spending rollout budget, and the error message can be fed back to the
proposer for a corrected typed diff.

\paragraph{Execution smoke tests.}
The second gate runs a small number of episodes against the candidate KB. In
ALFWorld, the verifier reloads KB-dependent modules so the newly written YAML is
actually used, runs \(n\) episodes with a fixed horizon, records success rate,
mean steps, fired rules, admissible-command misses, and compact failure traces.
In cold-start or sandboxed runs, this check is often run with a small \(n\) to
confirm that the KB is executable: rules can be loaded, predicates can be
evaluated, declared skills exist, emitted commands are admissible or recovered,
and traces are written. The benefit is fast feedback on integration failures,
such as a rule that names a nonexistent predicate, a skill binding that emits an
invalid command, or a KB cache that was not refreshed. Passing this smoke gate
only means the candidate is runnable; it does not establish the method's
claimed performance.

\paragraph{Focused validation.}
Focused validation is the first behavioral acceptance gate. It uses a task
family, seed window, or trajectory bank chosen to test the failure that
motivated the edit. For example, the \texttt{OpenGoalRecep} repair is evaluated
on closed-container deposit cases, because that is the behavior the edit is
supposed to repair. A process-tool edit is evaluated on clean/heat/cool tasks,
not only on generic pick/place tasks. The focused metric may be endpoint
success, or a predeclared trajectory-health metric when endpoint success is not
expected to change immediately. This prevents a broad or accidental change from
being accepted merely because it performs well on unrelated easy episodes.

\paragraph{Protected regression.}
Protected regression is the second behavioral acceptance gate. It reruns
previously solved task families or a broader fresh-seed window and rejects
the candidate if protected behavior regresses. In ALFWorld, after the
pick/place family is solved, later light-use or process-tool edits must not
break pick/place. In RoboSuite-style boundary studies, a candidate that improves
a focused nut-assembly window must also be checked against the broader task
set; if it only improves reward or an intermediate residual but fails the
environment success predicate, it is recorded as progress rather than accepted
policy knowledge.

\paragraph{Why smoke tests are still useful.}
Smoke tests are deliberately small because their role is engineering control,
not statistical evidence. They keep the edit loop cheap and debuggable: the
system can reject malformed YAML edits, stale-cache executions, missing skill
bindings, and impossible command emissions before running full verifier banks.
This makes the expensive checks more meaningful. Focused validation and
protected regression are then spent on candidates that are syntactically valid,
loadable, and executable. The separation also improves auditability: a rejected
candidate can be classified as a schema failure, execution failure, focused
failure, or regression failure, which is exactly the kind of localized
maintenance signal the white-box artifact is meant to provide.

\FloatBarrier

\section{Formal Properties of Verifier-Gated KB Editing}
\label{app:formal-properties}

This section gives the limited formal support behind the update rule in
Section~\ref{sec:updates}. The results are intentionally narrow. They prove
properties of the typed applier and verifier gate on declared finite validation
banks; they do not prove that Kintsugi will solve a benchmark or generalize
beyond the tested state distribution. Those broader claims are empirical and
are evaluated in Sections~\ref{sec:textagent-baselines}--\ref{sec:rq5-ablation}.

\paragraph{Objects and assumptions.}
Let \(\mathcal{K}\) be the set of well-typed executable KBs. A KB
\(K\in\mathcal{K}\) contains entries
\((\ell,k,\mathrm{type},\mathrm{content},\mathrm{prov})\) whose payloads
conform to the schema for layer \(\ell\), and whose executable references to
predicates, operators, skills, monitors, and goals are resolvable. Let
\(\mathcal{D}\) be the typed edit space. The deterministic applier is a partial
function
\[
  \mathrm{Apply}: \mathcal{K}\times\mathcal{D}
  \rightarrow \mathcal{K}\cup\{\bot\},
\]
where \(\bot\) denotes rejection before environment evaluation. The applier
returns a KB only after checking that the operation is admitted for the target
layer, the payload satisfies the operation schema, and all newly introduced or
modified executable references resolve.

For a finite trajectory or task bank \(B\), let \(M_B(K)\in[0,1]\) be the
deterministic success metric obtained by running the executor with KB \(K\) on
all episodes in \(B\), using the benchmark success signal. Some verifier gates
also include a predeclared trajectory-health metric \(H_B(K)\), such as
invalid-action rate or distance-to-goal residual, when endpoint success is not
expected to change immediately; when no such metric is declared, the health
metric branch below is disabled. For edit \(t\), \(T_t\) is the focused
validation bank and \(P_t\) is the protected regression bank. The acceptance
rule is:
\[
\begin{aligned}
\mathrm{Accept}_t(K,K')=1
\quad\Longleftrightarrow\quad&
K'\neq\bot,\;\; M_{P_t}(K')\ge M_{P_t}(K),\\
&\text{and either } M_{T_t}(K')>M_{T_t}(K)\\
&\text{or } \bigl(M_{T_t}(K')=M_{T_t}(K)
\;\wedge\; H_{T_t}(K')>H_{T_t}(K)\bigr).
\end{aligned}
\]
If the candidate is accepted, \(K_{t+1}=K'\); otherwise \(K_{t+1}=K_t\) and
the rejected hypothesis is recorded only as audit evidence.

\begin{proposition}[Type preservation]
\label{prop:type-preservation}
If \(K_t\in\mathcal{K}\) and
\(\mathrm{Apply}(K_t,\Delta_t)=K'\neq\bot\), then \(K'\in\mathcal{K}\).
\end{proposition}

\paragraph{Proof.}
The applier only returns a non-\(\bot\) value after validating the target layer,
operation, payload schema, and executable references introduced or modified by
\(\Delta_t\). Entries not touched by the edit remain well typed because
\(K_t\in\mathcal{K}\). Therefore every entry in \(K'\) satisfies its layer
schema and all executable references are resolvable, so \(K'\in\mathcal{K}\).

\begin{proposition}[Verifier-set monotonicity]
\label{prop:verifier-monotonicity}
For any accepted edit at iteration \(t\), the protected metric on the declared
regression bank does not decrease:
\[
  M_{P_t}(K_{t+1})\ge M_{P_t}(K_t).
\]
Moreover, the focused bank either improves in endpoint success or preserves
endpoint success while improving the predeclared trajectory-health metric.
\end{proposition}

\paragraph{Proof.}
For an accepted edit, \(K_{t+1}=K'\) and
\(\mathrm{Accept}_t(K_t,K')=1\). The three claims are exactly the three
behavioral conditions in the acceptance rule: protected success is
nondecreasing, and the focused bank either has strictly larger endpoint success
or equal endpoint success with strictly larger declared health metric.

\begin{proposition}[Rejected proposals do not change deployment]
\label{prop:rejected-proposals}
If a candidate edit is rejected by the applier or verifier, the deployed KB is
unchanged at that iteration: \(K_{t+1}=K_t\).
\end{proposition}

\paragraph{Proof.}
The update rule assigns \(K_{t+1}=K_t\) whenever
\(\mathrm{Apply}(K_t,\Delta_t)=\bot\) or
\(\mathrm{Accept}_t(K_t,\mathrm{Apply}(K_t,\Delta_t))=0\). Thus a rejected LLM
hypothesis can remain in the audit log, but it is not part of the executable
policy artifact and cannot affect zero-LLM deployment.

\paragraph{Scope of the formal claims.}
These propositions formalize the mechanism that makes Kintsugi conservative:
typed edits preserve the KB schema, accepted edits satisfy their declared
focused/protected verifier contracts, and rejected LLM hypotheses cannot
silently modify the policy. They are finite-bank guarantees. They do not
replace the empirical evidence in the main experiments, and they do not certify
behavior outside the state, perception, or skill interfaces covered by the
verifier banks.

\FloatBarrier

\section{Benchmark Protocol Details}
\label{app:benchmark-protocols}

This section collects the evaluation protocol details that are referenced from
the main experiment section. Keeping them here avoids overloading the headline
RQ tables with long descriptions of banks, action surfaces, and success
metrics.

\paragraph{Text-agent protocol.}
The paper-facing text-agent rows follow the Dual
Memory-aligned~\citep{wen2026aligning} evaluation contract used for
Table~\ref{tab:rq1-textagent}. ALFWorld uses the 134
\texttt{eval\_out\_of\_distribution}/valid-unseen games with a 50-step horizon.
The actor receives text observations and emits admissible text actions; success
is the environment \texttt{won} flag. WebShop uses 100 sessions with IDs
\(0\ldots99\), all products enabled, a 15-step budget, public text
observations, and only visible clickables or search actions. A WebShop episode
is counted as successful when reward \(r\ge0.99\), and the Score column reports
mean reward. TextCraft uses the 100 AgentGym AgentEval item IDs with a 40-action
budget over \texttt{get}, \texttt{craft}, and \texttt{inventory}; success is
defined by the environment \texttt{done} signal with positive reward.

The access-control rows use the same frozen accepted KBs but change the actor
interface. The no-KB row gives the language actor no KB context, the prompt row
renders the KB as context, and the tool row exposes typed KB retrieval. All LLM
access-control rows use \texttt{gpt-4o-2024-11-20} with temperature zero. The
symbolic-executor row runs the accepted KB directly and makes no test-time LLM
call.

\paragraph{Object-centric protocol.}
MetaWorld is evaluated on MT10 and MT50 with 50 rollouts per task, a 500-step
cap, and the environment binary success predicate. The held-out rows use
held-out reset seeds within the same task families rather than the ML10/ML45
held-out task-family protocol. Predicators uses four task families, 10 seeds,
and 50 evaluation tasks per family per seed, with success defined by symbolic
goal satisfaction under the planner/executor interface. RoboSuite is reported
as a single-arm boundary study: we use simulator-state observations, fixed
robot and controller settings, task-specific success predicates, a 500-step
cap, and five reset seeds for each of the 15 compatible single-arm tasks.
Focused RoboSuite repair windows are used only to localize accepted and
rejected edits; the headline quantitative row is the fixed-artifact single-arm
aggregate.

\paragraph{Scale and reporting.}
The paper reports episode counts rather than wall-clock time because simulator
backends and text-environment servers dominate runtime differently across
benchmarks. We use two WebShop reporting modes. The baseline-comparable
headline row uses 100 sessions because that is the bank used for the literature
comparison in Table~\ref{tab:rq1-textagent}. The full-bank/editor diagnostics
use 500 all-product sessions to test the accepted artifact and editor-model
construction behavior; these rows are not used as the direct baseline
comparison. The remaining appendix experiments use the same fixed-artifact
discipline: ALFWorld layer ablations run 120 episodes, MetaWorld uses 2500
episodes per split, Predicators uses 2000 total test tasks, and RoboSuite
reports \(68/75\) over the 15-task single-arm aggregate.

\paragraph{Repeated-run statistics.}
For Table~\ref{tab:rq1-textagent}, the Kintsugi and LLM access-control rows
report sample mean \(\pm\) standard deviation over five repeated runs under the
same declared task banks, budgets, and model setting. Literature baseline rows
are copied from their cited reports and are not rerun by us. For fixed-artifact
object-centric evaluations, we report the full declared denominators and seed
counts rather than a separate significance test.

\paragraph{Reproducibility artifacts and compute reporting.}
The supplemental artifact contains the implementation, accepted KBs, evaluation
scripts, and curated result manifests needed to reproduce the paper-facing
runs; upstream benchmark data are obtained from their original projects rather
than redistributed. The paper reports episode counts and protocol settings, but
does not fully standardize wall-clock runtime across text servers, simulator
backends, and API latency, so exact per-run hardware time should be treated as
artifact-level metadata rather than a paper-level claim.

Each agentic construction run writes an anonymized audit record. Proposal
budgets are computed from the following record schema:
\begin{quote}\small
\begin{verbatim}
proposal_attempt := record contains typed "diff"
apply_failed     := decision=="apply_failed" or apply_audit.error
accepted         := decision=="kept"
verifier_reject  := decision in {"reverted", "verifier_rejected"}
total_rejected   := apply_failed + verifier_reject
\end{verbatim}
\end{quote}
Each counted proposal therefore carries both a typed mutation and a gate
decision, summarized in Table~\ref{tab:app-kbdiff-budget-gates}.

\begin{table}[H]
\caption{\textbf{KBDiff proposal gates for budget accounting.}}
\label{tab:app-kbdiff-budget-gates}
\centering
\setlength{\tabcolsep}{3pt}
\renewcommand{\arraystretch}{1.05}
\begin{tabularx}{\linewidth}{@{}L{0.17\linewidth}L{0.23\linewidth}Y@{}}
\toprule
\textbf{Budget status} & \textbf{Gate} & \textbf{Evidence and artifact effect}\\
\midrule
Attempted & Proposal record &
Typed KBDiff names an operation, path, payload, evidence, and expected effect;
counted before acceptance.\\
Apply failed & Deterministic applier &
Invalid operation, path, payload, or unresolved predicate/skill reference; no
rollout claim is made and the KB is unchanged.\\
Verifier rejected & Focus/protect verifier &
Candidate applies, but focused success or declared health does not improve, or
the protected bank regresses; trace is retained as a rejected hypothesis and
the KB is restored.\\
Accepted & Commit gate &
Candidate applies, focused success or declared health improves, and protected
behavior is non-regressing; the candidate becomes the next frozen artifact.\\
Result-only & Incomplete audit &
Final benchmark metric may be reported, but proposal counts are excluded from
budget totals.\\
\bottomrule
\end{tabularx}
\end{table}

Table~\ref{tab:app-alfworld-gpt4o-budget} gives one complete numeric audit
for the weaker-editor ALFWorld construction run. The run starts from the empty
policy scaffold under the public ALFWorld adapter contract, the text-world
analogue of a robot manual, and uses the full 134-game valid-unseen bank at
horizon 50 for every verifier call.

\begin{table}[H]
\caption{\textbf{ALFWorld \texttt{gpt-4o} proposal-budget audit.}}
\label{tab:app-alfworld-gpt4o-budget}
\centering
\setlength{\tabcolsep}{4pt}
\renewcommand{\arraystretch}{1.08}
\begin{tabularx}{\linewidth}{@{}Xrrrrrrr@{}}
\toprule
\textbf{Run} & \textbf{Init.} & \textbf{Final} & \textbf{Prop.} &
\textbf{Acc.} & \textbf{Apply fail} & \textbf{Verif. rej.} &
\textbf{Eval eps.}\\
\midrule
ALFWorld \texttt{gpt-4o} &
\(0/134\) & \(134/134\) & 10 & 10 & 0 & 0 & 1474\\
\bottomrule
\end{tabularx}
\end{table}

The 10 accepted proposals consist of six \texttt{add\_task\_schema} edits
followed by four \texttt{add\_policy\_schema} edits. These edits do not provide
episode answers; they teach the artifact how to interpret public task families
and invoke the fixed action interface. The first nine edits close
verifier-declared contract gaps without changing endpoint success; after the
final policy schema is accepted, the frozen KB executes at \(134/134\). Thus the
budget counts include interface-health edits as accepted KBDiffs, not only edits
with immediate endpoint success gains.

For example, the ALFWorld \texttt{OpenGoalRecep} repair is accepted only after
the typed \texttt{add\_rule} payload applies, closed-container deposit cases
improve, and protected open-container/non-openable cases do not regress. By
contrast, RoboSuite \texttt{tool\_hang} frame-grip or wall-insert hypotheses
that improve intermediate geometry but leave the environment success predicate
false are verifier-rejected: the trace remains audit evidence, but the frozen
KB is restored.

For a budgeted run, the artifact manifest contains the environment/model
contract, final benchmark result, proposal-budget summary, anonymized audit
records, and final accepted KB. Rows without complete audit records are marked
as result-only and excluded from proposal-budget totals.

\FloatBarrier

\section{The Executable KB Outperforms Text-Agent Baselines}
\label{app:rq1-details}

This section expands Section~\ref{sec:textagent-baselines} and
Table~\ref{tab:rq1-textagent}. It asks whether the final accepted KB can serve
as a strong deployed policy on long-horizon text-agent tasks, while separating
endpoint task performance from the protocol details in
Section~\ref{app:benchmark-protocols}. The main comparison table reports
endpoint task performance, but the key deployment distinction is that Kintsugi
evaluates a frozen executable artifact with zero test-time LLM calls.

Baseline rows in the main RQ1 table are endpoint task-solving systems drawn
from the long-horizon language-agent literature. They are not required to share
Kintsugi's deployment discipline; instead, the comparison asks whether a frozen
executable policy artifact remains competitive against prompt-based,
reflection-based, memory-based, and state-tracking language agents. The
important distinction exposed by RQ1 is therefore both performance and policy
object: Kintsugi deploys the accepted KB itself, while language-agent baselines
typically continue to choose actions through a test-time language model.

The baselines cover several ways of making a language agent more persistent.
ReAct~\citep{yao2023react} is the prompt-agent foundation: it interleaves
reasoning traces with environment actions, so the LLM can update a plan while
interacting with ALFWorld or WebShop. Reflexion~\citep{shinn2023reflexion}
keeps the same test-time LLM decision loop but adds verbal reinforcement:
failed trajectories are reflected into text and stored as episodic memory for
future trials. ADaPT~\citep{prasad2024adapt} attacks long-horizon difficulty
through as-needed decomposition, recursively breaking a subtask down only when
the LLM executor cannot solve it directly. StateAct~\citep{rozanov2025stateact}
is a stronger base-agent prompt: it reinforces the goal at every step through
self-prompting and tracks evolving state through a chain-of-states.

The memory-oriented baselines move from single-trajectory prompting toward
cross-episode experience reuse. ExpeL~\citep{zhao2024expel} gathers training
experiences, extracts natural-language lessons, and recalls those insights at
inference without fine-tuning the model. AWM~\citep{wang2024agent} induces
reusable workflows from past agent trajectories and retrieves those workflows
to guide future action. Dual Memory~\citep{wen2026aligning}, the closest
paper-facing protocol match for our text-agent evaluation, explicitly separates
semantic progress guidance from symbolic feasibility verification: progress
memory keeps the agent aligned with the global objective, while feasibility
memory checks local constraints. WALL-E 2.0~\citep{zhou2025wall} is closest to
Kintsugi in its use of symbolic environment knowledge: it aligns an LLM-based
world model with extracted action rules, knowledge graphs, and scene graphs,
then uses the LLM in a model-predictive planning loop.

These are strong baselines because they cover the main prior strategies:
interleaved reasoning/action, verbal reflection, decomposition, state tracking,
experiential memory, workflow memory, dual progress/feasibility memory, and
neurosymbolic world-model alignment. Their shared limitation for our claim is
where final action selection lives. In all of them, the LLM remains an active
test-time actor or planner. Even when memory or symbolic checks are available,
the model must retrieve the right information, bind it to the current state,
decide which constraint matters, and emit a valid action at each step.

Kintsugi changes that boundary by compiling accepted knowledge into an
executable KB. At test time, the symbolic executor reads the environment state,
evaluates typed predicates and policies, and emits the action directly. The LLM
is used only between rollouts to propose candidate edits, and a candidate
becomes policy only after deterministic application and validation.

\paragraph{RQ1 run records and editor-context diagnostics.}
The headline RQ1 table should be read as the baseline-comparable endpoint
comparison under Section~\ref{app:benchmark-protocols}. The accepted artifacts
behind those rows were authored by the stronger \texttt{gpt-5.5} editor and
then executed as frozen KBs with zero test-time LLM calls. Separately, full-bank
and \texttt{gpt-4o} diagnostics test robustness and weaker-editor construction
behavior before deployment. The distinction matters: endpoint rows compare
frozen policies to baselines, whereas editor-context diagnostics evaluate
proposal quality and convergence of the editor loop.

\begin{table}[H]
\caption{\textbf{Key RQ1 construction records.}}
\label{tab:app-rq1-current-records}
\centering
\setlength{\tabcolsep}{8pt}
\renewcommand{\arraystretch}{1.08}
\begin{tabular}{@{}lcc@{}}
\toprule
\textbf{Benchmark} &
\makecell{\textbf{\texttt{gpt-5.5}}\\\textbf{SR / score}} &
\makecell{\textbf{\texttt{gpt-4o}}\\\textbf{SR / score}}\\
\midrule
ALFWorld & \(134/134=100.0\%\) & \(134/134=100.0\%\)\\
WebShop & \(290/500=58.0\%\; /\;0.7863\) & \(283/500=56.6\%\; /\;0.7620\)\\
TextCraft & \(100/100=100.0\%\) & \(96/100=96.0\%\)\\
\bottomrule
\end{tabular}
\end{table}

The \texttt{gpt-5.5} columns report the stronger-editor accepted artifacts, deployed
with zero test-time LLM calls. The \texttt{gpt-4o} columns report weaker-editor
diagnostics:
ALFWorld is from-zero under the public policy-schema contract, WebShop reports
the extended 500-session all-product diagnostic for the accepted artifact, and
TextCraft is a from-zero run. This WebShop row is therefore separate from the
100-session Dual Memory-aligned headline value in Table~\ref{tab:rq1-textagent}.
These records show that the agentic method still works with a weaker editor,
but the weaker editor benefits more from explicit schema/interface guidance and
shows less stable construction quality.

The \(100\%\) ALFWorld and TextCraft rows should be read as closure on their
declared evaluation banks, not as open-ended generalization. Their credibility
comes from repeated fixed-artifact execution, ALFWorld task-type breakdowns,
access-control comparisons, and layer ablations. The reason ceiling performance
is attainable is structural: ALFWorld valid-unseen tasks and TextCraft recipes
have recurring symbolic schemas that can be represented as executable
preconditions, dependencies, monitors, and recovery rules.

WebShop should also be interpreted as the least natural fit for the current
artifact design. Kintsugi is strongest when failures decompose into stable
object-centric planning contracts: predicates, preconditions, tool use,
subgoal order, monitors, and recovery. WebShop instead stresses retrieval and
ranking under incomplete public snippets, hidden decisive attributes, and fuzzy
reward matching. The KB can encode query, constraint, and option-binding rules,
but it does not control the catalog evidence exposed by search. Therefore the
WebShop rows are evidence of a useful but less aligned application, not a clean
editor-strength comparison.

The ALFWorld editor diagnostic is particularly informative for the claim
boundary. In the stronger-editor ALFWorld run, \texttt{gpt-5.5} reaches
\(134/134\) after three accepted repair rounds, adding 13 rule-level edits. This
is the cleaner from-zero construction evidence used by the main cold-start
ledger. In contrast, without the public high-level policy-schema contract, the
\texttt{gpt-4o} editor repeatedly proposed malformed or overly local low-level
rules and did not converge. With the public environment contract, it added six
task-family schemas and four policy schemas, reaching \(134/134\) after ten
iterations. The final \texttt{gpt-4o} KB contains no learned
\texttt{procedural.rules} table and is solved by compiling the accepted policy
schemas through the deterministic executor. This supports the intended Kintsugi
claim: stronger editors can infer and optimize policy structure faster, while a
weaker editor can still construct a verified executable artifact when the
environment interface is explicit. In both cases, the deployed policy is the
frozen accepted artifact rather than a test-time LLM.

\paragraph{How symbolic planning is executed.}
At deployment, Kintsugi does not ask the LLM to write a plan. The executor
first maps the benchmark observation and goal into the KB's typed state. In
ALFWorld, this means extracting the target object, required receptacle or tool,
inventory state, visible containers, open/closed state, and admissible text
actions. In TextCraft, it means reading the requested item, current inventory,
known recipes, and available \texttt{get}/\texttt{craft} actions. In WebShop,
it means tracking query terms, visible product candidates, options, prices, and
hard constraints. The KB then supplies task schemas, preconditions, operator
effects, monitors, and recovery rules over this state. Planning is therefore a
typed policy evaluation: select the active goal schema, choose the first
unsatisfied subgoal, test operator preconditions against the current state, and
emit the grounded action whose effect advances that subgoal.

The mapping from KB to action is different across the three text benchmarks
but has the same structure. In ALFWorld, a goal such as cleaning, heating, or
cooling an object maps to an ordered schema: locate the object, pick it up,
move to the correct tool, open the tool if needed, place the object, activate
or use the tool, retrieve the object, and satisfy the final receptacle goal.
Recovery monitors handle common deviations, such as a closed receptacle, an
object not found in the first searched container, or a processed object that
must be revisited after a tool action. In TextCraft, the target item maps to a
recipe dependency graph; the executor repeatedly selects the missing leaf
ingredient or craftable intermediate until the final recipe is executable. In
WebShop, the policy maintains hard constraints and candidate scores before
purchase, but the endpoint is harder because the public observation may not
expose the decisive catalog evidence needed for a perfect ranking decision.

\paragraph{How agentic repair changes the endpoint policy.}
During training, failed rollouts are converted into localized KB hypotheses.
A failure trace is not treated as a general instruction to ``try harder.'' It
is assigned to a layer: missing grounding, missing predicate, missing operator
effect, missing policy schema, missing monitor, or missing recovery. The
proposer can then suggest a typed KBDiff, such as adding a process-tool schema,
adding a revisit rule after microwave/fridge/sink use, or adding a recipe
dependency guard. The verifier accepts the edit only if it improves the
focused failure family and does not regress protected solved cases. This is
why successful endpoint behavior is cumulative: later policies inherit earlier
accepted schemas and add only the missing transition.

Failures that are not accepted are equally important for interpreting RQ1.
If an edit improves an intermediate trace property but fails the benchmark
success predicate, it remains a rejected hypothesis. In ALFWorld and TextCraft,
the remaining benchmark failures were closed because the residual modes could
be written as stable symbolic contracts. In WebShop, some residual modes cannot
be closed by another local symbolic rule: search ranking, missing public
attributes, and option ambiguity can prevent the executor from seeing enough
evidence to make a guaranteed purchase. This is why the final WebShop result
improves over baselines but does not reach the same ceiling as ALFWorld and
TextCraft.

This explains why the largest gains appear on ALFWorld and TextCraft. Both
benchmarks have recurring symbolic structure: household process schemas in
ALFWorld and recipe/inventory dependency closure in TextCraft. Once those
schemas are represented as executable policy rules, the executor does not need
to re-derive them in context. WebShop is less closed because product retrieval
and ranking depend on public snippets and search ordering; this is why the
margin over Dual Memory~\citep{wen2026aligning} is smaller there and why the
KB-tool language actor can underperform the rendered-prompt actor. The
important RQ1 result is therefore not only that the final numbers are higher,
but that the strongest policy is the artifact itself rather than a language
model interpreting the artifact.

\paragraph{Why the endpoint reaches \(100\%\) on ALFWorld and TextCraft.}
The endpoint can reach the ceiling when the remaining failure modes are
expressible as stable symbolic contracts. ALFWorld has long horizons, but its
valid-unseen tasks are built from recurring household schemas: search, pickup,
open, place, light use, and process-tool variants. After the accepted rules
cover those schemas and their revisit/recovery cases, every task type in the
134-game bank has an executable proof path. TextCraft is even more
structure-preserving: once recipes, inventory checks, and prerequisite closure
are represented as a dependency policy, each of the 100 evaluation items can be
solved by executing the graph. WebShop does not have the same closure property
under public text observations, because product retrieval and ranking can hide
the decisive evidence. This is why WebShop remains below \(100\%\) even though
the KB improves the policy.

\FloatBarrier

\section{Capability Resides in the Executable KB Artifact}
\label{app:rq2-details}

This section expands Section~\ref{sec:rq2-artifact}. It treats the accepted KB
as a memory and policy artifact, asking whether the KB stores task knowledge in
a form that is useful to an LLM actor, and whether that same knowledge is more
reliable when executed directly rather than reinterpreted by an LLM at every
step. The cold-start ledger records where this memory comes from: policy
knowledge enters the artifact only through verifier-accepted typed edits. The
access-control analysis then freezes the completed KB and changes only how the
actor accesses that memory.
Table~\ref{tab:app-self-evolution} reports cold-start evolution using only the
minimal ledger fields. In ALFWorld, the accepted rounds add pick/place,
light-toggle, and process-tool/revisit repairs; in MetaWorld, one accepted
round installs the predicates, skills, and rules for a pick-place family.
The access-control rows themselves are already reported in
Table~\ref{tab:rq1-textagent}; here we analyze their failure signatures and add
the ALFWorld task-type diagnostic in Table~\ref{tab:app-alfworld-breakdown}.

\begin{table}[H]
\caption{\textbf{Cold-start artifact evolution.}}
\label{tab:app-self-evolution}
\centering
\setlength{\tabcolsep}{6pt}
\renewcommand{\arraystretch}{1.08}
\begin{tabular}{@{}llcc@{}}
\toprule
\textbf{Benchmark} & \textbf{Artifact} & \textbf{Edits} & \textbf{Success}\\
\midrule
ALFWorld & Scaffold & 0 & \(0/134 = 0.0\%\)\\
ALFWorld & Iter 1 & +8 & \(41/134 = 30.6\%\)\\
ALFWorld & Iter 2 & +2 & \(59/134 = 44.0\%\)\\
ALFWorld & Iter 3 & +3 & \(\mathbf{134/134 = 100.0\%}\)\\
MW pick-place & Scaffold & 0 & \(0/50 = 0.0\%\)\\
MW pick-place & Iter 1 & +27 & \(\mathbf{50/50 = 100.0\%}\)\\
\bottomrule
\end{tabular}
\end{table}

The cold-start ledger establishes acquisition: policy knowledge enters the
artifact only through verifier-accepted edits. The table is intentionally
coarse: it records when an executable artifact becomes stronger, not every
rejected hypothesis considered by the editor. This keeps the acquisition
evidence separate from the access-control evidence in the main table, where the
same completed artifact is exposed through different interfaces.

The ALFWorld row above is the \texttt{gpt-5.5} rule-level cold-start ledger
used for the local-editability analysis. The \texttt{gpt-4o} policy-schema
diagnostic in Table~\ref{tab:app-rq1-current-records} uses a different mutation
surface: it accepts six \texttt{add\_task\_schema} edits followed by four
\texttt{add\_policy\_schema} edits, then executes the accepted schemas through
the fixed ALFWorld executor. These two ledgers answer different questions. The
stronger-editor ledger shows faster from-zero construction through localized
symbolic repairs; the weaker-editor ledger shows that the agentic loop still
works when the public environment interface makes the policy-schema targets
explicit.

The access-control rows in Table~\ref{tab:rq1-textagent} are designed to
separate four mechanisms that are otherwise easy to conflate. The LLM w/o KB
row is a step-by-step language actor under the same task protocol. It is
therefore closest to the action-selection burden faced by prompt agents such
as ReAct~\citep{yao2023react}: the model must infer the goal, remember the
state, infer preconditions, and choose an admissible action from text alone.
Its failures are mostly not single-step language misunderstandings. They are
long-horizon control failures: losing the target object after moving through
rooms, applying a process action before the object is in the right tool,
forgetting an intermediate TextCraft ingredient, or buying a WebShop item that
matches a surface keyword but violates a hidden hard constraint.

The KB-prompt row asks a different question: if the learned artifact is
rendered as natural-language or structured prompt context, can a test-time LLM
use it as advice? This improves over no KB because the prompt exposes task
schemas, affordances, preconditions, recipe dependencies, and common recovery
patterns. In ALFWorld, the LLM can read that heat/cool/clean goals require
tool-specific transport and revisit behavior; in TextCraft, it can read recipe
dependencies and inventory checks; in WebShop, it can read constraint-tracking
and purchase-gating rules. However, prompt access does not make the KB the
policy. The LLM still has to bind the rendered rule to the current observation,
decide whether the rule applies now, and emit the correct action. The prompt
row therefore measures whether the artifact is informative to an LLM, not
whether it is fully executed.

The KB-tool row tests a more structured interface. Instead of giving the whole
artifact as prompt text, the LLM can retrieve typed facts from the accepted KB:
object categories and affordances, symbolic preconditions, operator effects,
goal schemas, recipe dependencies, monitors, and recovery suggestions. This is
why tools are much stronger on ALFWorld and TextCraft. These two domains have
compact symbolic state and discrete action grammars, so a retrieved rule often
maps almost directly to the next action. For example, retrieving the current
ALFWorld process schema tells the actor which object, receptacle, and tool are
load-bearing; retrieving the TextCraft dependency graph tells the actor which
ingredient is still missing. In this regime, KB tools reduce hallucinated
preconditions and shorten the LLM's search over possible next actions.

The tool row is still not the deployed Kintsugi policy. It is an LLM-mediated
policy that calls into the KB. This distinction matters for robotics and
planning, which are the main target setting for the artifact design. A robot
planner cannot treat a rule such as ``insert after stable contact'' as a vague
hint; it needs a state predicate, a precondition, an operator or skill binding,
a monitor, and a success/failure condition. Kintsugi's symbolic executor owns
that contract. It evaluates the predicates and policy schemas itself, triggers
monitors and recoveries deterministically, and emits an action without asking
the LLM to reinterpret the retrieved fact. Thus the executor row tests whether
the KB is a policy object, while the prompt/tool rows test whether the same
policy object is useful when exposed to a language actor.

The WebShop exception shows the boundary of the current tool interface rather
than a contradiction of the KB claim. WebShop product selection is a fuzzy
ranking problem over public snippets, option strings, prices, and search
ordering. The current KB tool returns local symbolic or state facts; it does
not own the full retrieval-and-ranking loop, does not see hidden catalog
evidence, and can fragment a comparison that is easier for an LLM to perform
when the KB is rendered in one prompt. A tool call can also make the actor
over-focus on a retrieved constraint while missing a better candidate in the
visible product list. This is why KB prompt access can outperform KB tools on
WebShop even though both improve over no-KB acting. For embodied planning
domains such as ALFWorld, TextCraft, MetaWorld, and RoboSuite-style symbolic
control, the useful KB components are local and executable; for WebShop, the
missing component is a KB-owned retrieval/ranking policy rather than another
local precondition fact.

\begin{table}[H]
\caption{\textbf{ALFWorld access-control breakdown by task type.}}
\label{tab:app-alfworld-breakdown}
\centering
\setlength{\tabcolsep}{4pt}
\renewcommand{\arraystretch}{1.06}
\begin{tabular}{@{}lrrrrr@{}}
\toprule
\textbf{Type} & \textbf{N} & \textbf{w/o KB} & \textbf{KB prompt} &
\textbf{KB tools} & \textbf{Exec.}\\
\midrule
Pick & 24 & 16/24 & 20/24 & 24/24 & 24/24\\
Light & 18 & 4/18 & 6/18 & 18/18 & 18/18\\
Clean & 31 & 8/31 & 17/31 & 31/31 & 31/31\\
Heat & 23 & 5/23 & 8/23 & 21/23 & 23/23\\
Cool & 21 & 7/21 & 8/21 & 21/21 & 21/21\\
Two-obj. & 17 & 7/17 & 9/17 & 17/17 & 17/17\\
\midrule
\textbf{Total} & \textbf{134} & \textbf{47/134} & \textbf{68/134} &
\textbf{132/134} & \textbf{134/134}\\
\bottomrule
\end{tabular}
\end{table}

The task-type breakdown shows where access pattern matters. Rendering the KB
as prompt context helps some process tasks but still leaves action choice to
the language model. Typed retrieval is much closer to execution, but the model
must still compose retrieved facts into an action. Executing the artifact
directly removes that interpretation step and solves every held-out task type.
The point is not that the retrieved KB is uninformative: the near-perfect
LLM w/ KB tools row shows that the artifact contains the needed knowledge. The
remaining gap comes from using a language actor to interpret that knowledge
instead of letting the symbolic executor run the accepted rules directly.
\paragraph{Text-agent convergence and case analysis.}
The access-control rows should be read together with the edit history. The
three LLM rows do not create new policy knowledge; they only expose the same
frozen KB as no context, rendered prompt context, or typed retrieval. The
symbolic row is different: the accepted KB itself is the deployed policy. This
is the main reason that the symbolic executor can be stronger than an LLM that
has access to the same information.

In ALFWorld, the KB changes the policy from a generic text-action loop into a
stateful household-task executor. The cold-start scaffold solves \(0/134\)
games because it has observation and action contracts but no task schemas.
The first accepted round adds the basic search, pickup, open-if-needed, and
place chain, moving the artifact to \(41/134\). The second round adds the
light-toggle schema, reaching \(59/134\). The third round adds process-tool and
revisit behavior for clean, heat, cool, and related container cases, reaching
\(134/134\). A typical successful transformation is a heat task: the policy no
longer treats ``heat the potato'' as one semantic hint, but decomposes it into
locating the potato, carrying it to the microwave, opening the microwave,
placing the potato inside, activating the microwave, retrieving the object, and
checking the goal. The symbolic policy changes because these steps become typed
rules and monitors, not prompt text. When the LLM rows fail, the trace usually
shows a plausible admissible action chosen at the wrong time, such as searching
the wrong container after the correct object-location rule is already available.
Thus the residual ALFWorld gap is mostly an interpretation gap between KB
knowledge and LLM action choice, not missing executable knowledge in the KB.

In TextCraft, the useful KB change is a different kind of symbolic policy
rewrite. The accepted artifact turns recipes and inventory observations into a
dependency graph over \texttt{get}, \texttt{craft}, and \texttt{inventory}
actions. A successful case is a target item whose recipe requires several
intermediate materials: the executor plans backward from the requested item,
checks whether each ingredient is already present, collects only missing
materials, and crafts prerequisites before the final item. This directly fixes
the dominant LLM failure pattern, where the actor crafts too early, repeats a
collection action after the inventory is sufficient, or forgets an intermediate
recipe. The \(97/100\) KB-tool result shows that the exposed rules are already
almost sufficient for a language actor, but the \(100/100\) symbolic result
comes from executing the dependency policy rather than reinterpreting it at
each step.

WebShop is the important counterexample and is therefore reported without
smoothing. The KB does improve the task by maintaining query terms, hard
constraints, option choices, price/attribute scores, and purchase gating under
the 100 Dual Memory-aligned~\citep{wen2026aligning} sessions. A successful
WebShop case is one where the symbolic policy keeps the requested attributes
separate from soft textual
similarity and refuses to buy a product whose visible option or price violates
a hard constraint. The remaining failures have a different character from
ALFWorld and TextCraft. Public product snippets do not always expose the
decisive attribute, search ranking can bury a valid product behind near misses,
and the current typed tool returns local KB/state facts rather than owning the
full retrieval-and-ranking loop. This explains why rendered KB prompt access
can beat the tool interface on WebShop: the prompt gives the language model a
single context for fuzzy product comparison, whereas the tool interface can
fragment that comparison across calls. In this setting convergence stops when
new symbolic edits improve a visible constraint case but do not reliably
improve final reward under the public observation interface. We therefore treat
the WebShop tool result as evidence about an interface boundary, not as a
failure of the KB to contain useful shopping knowledge.

\section{Localized KB Edits Repair Target Behaviors}
\label{app:rq3-details}

This section expands Section~\ref{sec:localized-kb-edits} and
Table~\ref{tab:rq3-editability}. It asks whether policy behavior can be changed
through localized typed KB edits without off-target regression, naming the edit
loci and validation structure behind each accepted round. The ALFWorld
cold-start ledger gives a controlled case because each accepted edit round
targets a named family of rules and predicates, and each commit is checked
against behavior that was already solved.

The three accepted ALFWorld rounds have different edit loci. The first round
repairs the basic pick/place chain: the executor learns how to search for a
target object, take it, navigate to the requested receptacle, open the
receptacle if needed, and deposit the object. This moves the targeted
pick/place family from \(0/41\) to \(41/41\). The second round adds the
light-use transition for \texttt{look\_at\_obj\_in\_light}: once the target is
held, the policy must locate a lamp and issue a \texttt{use} command rather
than deposit the object. This repairs \(0/18\) to \(18/18\) while preserving the
already solved pick/place behavior. The third round adds the process-tool
chains for \texttt{clean}, \texttt{heat}, and \texttt{cool}; these repairs add
tool navigation, tool-specific process actions, and the revisit behavior needed
after processing. The target family moves from \(0/75\) to \(75/75\), again
with no protected regression in the previously solved families.

The important point is not only that success increases. Each accepted update
has a bounded owner. Missing deposit behavior belongs to policy schemas and
openability predicates; missing lamp behavior belongs to the task schema and
toggle-use binding; missing clean/heat/cool behavior belongs to process-tool
semantics, process predicates, and skill bindings. This is the white-box
maintenance claim: a failure trace does not become a generic training example,
but a localized edit hypothesis with a focused validation set and a protected
regression set.

\paragraph{Interactive white-box repair loop.}
The RQ3 experiment is interactive in the sense that each round exposes the
current symbolic failure before changing the policy. A rollout stores the
observation, parsed state, active goal schema, predicates that evaluated true
or false, the selected rule, the emitted action, and the environment response.
When a trajectory fails, the analysis step does not ask for an unconstrained
new policy. It asks which symbolic contract was missing or wrong. The answer is
then represented as a typed KBDiff against a named layer: grounding, predicate,
ontology/affordance, operator effect, policy schema, monitor, recovery, or goal
template. The edit is applied to the KB, loaded by the same executor, and
checked by smoke tests, a focused validation bank for the target failure, and a
protected regression bank for behavior already solved.

This is why we call the RQ3 evidence white-box rather than just iterative
prompt improvement. The object being changed is visible: a rule, predicate,
operator, or schema in the KB. The reason for the change is also visible:
the trace shows which predicate was missing, which precondition blocked an
operator, or which rule fired too early. After the edit commits, the symbolic
executor's behavior changes deterministically because its rule set has changed;
there is no hidden parameter update and no test-time prompt memory that must be
reinterpreted by an LLM.

\paragraph{What changes in the symbolic policy.}
The first ALFWorld round turns the scaffold into a basic object-transfer
policy. Before the edit, the executor can read observations and issue actions,
but it does not own a complete schema for finding an object and depositing it
at a goal receptacle. The accepted KBDiff adds search/pick/place policy rules,
open-if-needed behavior, and the predicates needed to distinguish held object,
current receptacle, openable receptacle, and completed placement. Symbolically,
the policy changes from a fallback action loop into a schema with subgoals:
find target, acquire target, navigate to goal receptacle, satisfy receptacle
preconditions, and deposit. This is why the pick/place family moves from
\(0/41\) to \(41/41\).

The second round is a smaller but cleaner locality test. The already accepted
pick/place rules are not rewritten. The failure trace for
\texttt{look\_at\_obj\_in\_light} shows that holding the target object is not
enough: the goal requires a lamp interaction rather than a receptacle deposit.
The accepted edit adds a light-use schema and a toggle/use binding. The
symbolic executor now recognizes a lamp as the relevant tool object, navigates
to it, and emits \texttt{use} at the correct phase. Because the edit is scoped
to the light goal schema, the earlier pick/place bank remains protected and
does not regress.

The third round adds the process-tool layer needed for clean, heat, and cool
tasks. Here the KB changes are more structural: the goal parser maps the task
to a process type, ontology entries bind process types to tools such as sink,
microwave, and fridge, policy schemas add transport-into-tool and retrieve-
after-process phases, and monitors check whether the object must be revisited
after the tool action. This changes the executor from a pure object-transfer
policy into a process-aware household policy. The repair succeeds because the
same symbolic pattern covers many surface goals while still using specific
tool bindings and preconditions for each process.

\paragraph{Why rejected edits matter.}
Several plausible edits are not accepted because they are not local enough.
A broad rule such as always opening containers can help closed-receptacle
failures but can waste steps or issue invalid actions on non-openable targets.
A generic process fallback can appear to make progress on heat/cool/clean
traces while failing to retrieve the processed object or breaking already
solved pick/place tasks. A rule that fires whenever an object is held can
interfere with light tasks and two-object tasks. These candidates remain useful
diagnostics, but they do not become policy. The verifier requires the target
failure family to improve and the protected family to stay fixed; otherwise the
KBDiff is rejected and the deployed KB is unchanged.

\paragraph{Why this leads to success rather than drift.}
The successful trajectory is cumulative because each accepted edit tightens a
specific symbolic contract. Round 1 installs object transfer. Round 2 adds a
separate light-use branch. Round 3 adds process-tool branches and revisit
behavior. Later rules do not replace earlier rules; they add missing branches
or preconditions under named goal schemas. This prevents the common failure of
interactive prompt repair, where a new natural-language instruction improves
one case but changes the model's behavior elsewhere. In Kintsugi, the edit
target, rule priority, trigger predicate, and validation bank are explicit, so
success comes from accumulating verified local repairs rather than from
implicitly steering an LLM with a longer prompt.

\paragraph{Example repair.}
One early ALFWorld failure mode occurs after the agent has found and picked up
the target object and navigated to the goal receptacle. The trace shows that the
current receptacle is an openable container, the target is held, and the
episode is ready for deposit, but the container is still closed. A prompt-only
agent might record this as a verbal reminder such as ``remember to open
containers.'' Kintsugi instead localizes the repair to the policy-schema and
predicate interface: the missing condition is
\texttt{current\_openable\_closed}, and the missing action is an
\texttt{OPEN} step before \texttt{PUT}. The accepted repair is the
\texttt{OpenGoalRecep} rule described above. Its focused validation set
contains closed-container deposit cases, while the protected set contains the
already solved open-container and non-openable receptacle cases. The repair
commits only if the former improves and the latter does not regress.

\begin{table}[t]
\centering
\caption{\textbf{Accepted-edit ledger summary.}}
\label{tab:app-ledger}
\setlength{\tabcolsep}{3pt}
\renewcommand{\arraystretch}{1.08}

\begin{tabular*}{0.75\textwidth}{@{\extracolsep{\fill}}lccc@{}}
\toprule
\textbf{Run} & \textbf{Initial} & \textbf{Edits} & \textbf{Final}\\
\midrule
ALFWorld cold-start & 0/134 & 13 & 134/134\\
MW pick-place & 0/50 & 27 & 50/50\\
Pred. cover & Scaffold & 1 & 500/500\\
Pred. blocks & Scaffold & 1 & 500/500\\
Pred. painting & Scaffold & 1 & 500/500\\
Pred. pybullet-cover & Scaffold & 1 & 497/500\\
\bottomrule
\end{tabular*}

\end{table}

Rejected hypotheses remain in the audit log and are not counted as policy
updates. The final column reports only verifier-accepted artifacts. This
distinction is important for the editability claim: a generated hypothesis is
not a policy update until it has passed typed application, focused validation,
and protected regression. The ledger therefore records both the small number of
committed updates and the fact that failed hypotheses remain auditable rather
than silently changing the deployed policy.

\FloatBarrier

\section{KB-Driven Policies Extend to Object-Centric Manipulation}
\label{app:rq4-details}

\begin{wraptable}{r}{0.52\textwidth}
\vspace{-8pt}
\caption{\textbf{Reuse and boundary evidence.}}
\label{tab:app-reuse-boundary}
\centering
\setlength{\tabcolsep}{3pt}
\renewcommand{\arraystretch}{1.08}
\begin{tabular*}{0.52\textwidth}{@{\extracolsep{\fill}}lc@{}}
\toprule
\textbf{Setting} & \textbf{Result}\\
\midrule
MT-50 train & 2466/2500 = 98.6\%\\
MT-50 held-out & 2458/2500 = 98.3\%\\
Predicators & 1997/2000 = 99.85\%\\
RoboSuite single-arm & 68/75 = 90.7\%\\
\bottomrule
\end{tabular*}
\vspace{-8pt}
\end{wraptable}

This section expands Section~\ref{sec:rq4-object-centric}. It tests whether
the same KB-driven symbolic-policy interface extends beyond text-agent
environments when the substrate exposes object-centric state and calibrated
action or skill bindings, separating mature object-centric reuse from
transitional RoboSuite boundary evidence. The rows in
Table~\ref{tab:app-reuse-boundary} deliberately mix different evidentiary
regimes, so the interpretation belongs in prose rather than inside the table.
MetaWorld and Predicators evaluate mature object-centric KBs, and RoboSuite is
a transitional single-arm boundary study rather than a full visual
robot-learning claim.

The MetaWorld train and held-out rows differ only slightly, suggesting that the
KB is not merely memorizing reset seeds. The Predicators row tests the same
symbolic-policy idea in external TAMP-style domains. The RoboSuite row is
stricter than symbolic plan validity because success is measured under true
policy-action rollouts, but it uses simulator state and is therefore reported
as boundary evidence.

\paragraph{Object-centric convergence summary.}
Across the non-text experiments, the symbolic policy changes in the same broad
way: it stops acting as a flat collection of task names and starts acting as an
object-centric controller over roles, relations, phase predicates, monitors,
and skill bindings. In MetaWorld, this means the same KB can reuse object roles
and contact/support relations across all 50 tasks while swapping in
task-specific phase schemas. In Predicators, it means goal atoms are rewritten
into explicit symbolic preconditions such as which block must be covered, which
table height makes an \texttt{OnTable} fact true, or why a painted object must
be washed and dried before being boxed. In RoboSuite, it means rollout failures
are localized to the boundary between a symbolic intent and a continuous skill:
yaw-aware transport, nut-contact insertion, scene-conditioned ordering, and
recovery monitors become explicit policy knowledge only after they improve
focused validation without breaking protected checks.

The remaining failures are therefore not all the same kind of miss. MetaWorld
residuals mostly reflect precision and contact-timing families, where another
high-level symbolic rule is not accepted unless it improves those difficult
episodes without weakening the many already-solved tasks. Predicators has only
three misses, all in the embodied \texttt{pybullet\_cover} setting; the
abstract cover domain is solved by the same task-level KB, so these misses are
treated as IK/controller variance rather than absent cover knowledge. RoboSuite
is stricter still: \texttt{tool\_hang}, the remaining \texttt{nut\_assembly}
seed, and the door miss require contact-preserving insertion, orientation
channels, and validated low-level recovery monitors. These are precisely the
cases where the loop can diagnose progress but cannot safely accept a policy
update until the continuous interface exposes a reliable contract.

Table~\ref{tab:app-robosuite-aggregate} expands the RoboSuite row used in
Table~\ref{tab:rq4-object-centric}. The quantitative claim uses the
15 compatible single-arm tasks with five reset seeds each.

\begin{table}[H]
\caption{\textbf{RoboSuite single-arm breakdown.}}
\label{tab:app-robosuite-aggregate}
\centering
\setlength{\tabcolsep}{8pt}
\renewcommand{\arraystretch}{1.05}
\begin{tabular}{@{}lc@{}}
\toprule
\textbf{Task} & \textbf{Success}\\
\midrule
Aggregate & 68/75 = 90.7\%\\
\midrule
door & 4/5\\
lift & 5/5\\
nut\_assembly & 4/5\\
nut\_assembly\_round & 5/5\\
nut\_assembly\_single & 5/5\\
nut\_assembly\_square & 5/5\\
pickplace\_bread & 5/5\\
pickplace\_can & 5/5\\
pickplace\_cereal & 5/5\\
pickplace\_milk & 5/5\\
pickplace\_multi & 5/5\\
pickplace\_single & 5/5\\
stack & 5/5\\
tool\_hang & 0/5\\
wipe & 5/5\\
\bottomrule
\end{tabular}
\end{table}

\paragraph{Why the remaining single-arm failures persist.}
The \(68/75\) RoboSuite result should not be read as an iteration that simply
stopped before trying enough patches. The agentic loop generated both accepted
and rejected hypotheses, and the verifier admitted a candidate only when it
improved the focused window without violating the environment success predicate
or protected regression checks. Several rejected edits improved an intermediate
residual, added extra recovery actions, or solved one seed while breaking
another. These rejected hypotheses are useful evidence: they show where the
symbolic policy is already choosing the right high-level intent, but the
grounding, recovery, or skill-interface contract below that intent is still
under-specified.

The clearest remaining case is \texttt{tool\_hang}, the only single-arm task
in this aggregate that remains at \(0/5\). Bootstrap and frame-grip probes
could move the frame and improve intermediate relations such as
\texttt{frame\_tip\_close\_to\_base}, but they did not make
\texttt{frame\_rod\_between\_walls} or \texttt{tool\_on\_frame} true. Some
waypoint variants improved the apparent insertion geometry, while other
variants pushed the frame back out of the insertion window. Because these
edits failed the final environment success predicate, the verifier recorded
them as progress evidence rather than solved policy knowledge. The failure is
therefore not well explained as missing high-level task order. The missing
contract is a real frame-pivot / stand-wall insertion and hook-placement
schema, with explicit contact-preserving translation, orientation channels,
and success monitors. The loop can say what is missing, but it cannot safely
commit a KB rule until the action interface exposes a repeatable way to keep
the frame inside the narrow contact window while placing the tool.

The remaining \texttt{nut\_assembly} failures have a different structure.
Accepted edits already repair several subcases: RoundNut contact, SquareNut
handle-first grounding, and scene-conditioned multi-nut ordering. Some focused
windows reached \(5/5\), while an older strict regression window stayed at
\(4/5\). The rejected candidates are informative here as well. Tighter contact
windows, blind subgoal resets, generic settling, and low-level transport
recoveries either did not improve the paired strict window, slowed successful
trajectories, or solved one seed while breaking another. This points to a
contact/insertion and scene-conditioned sequencing boundary. The next safe
edit is not a looser success threshold, but a stronger KB-owned verification
contract for when a nut is actually controlled, aligned, inserted, and safe to
advance past.

The \texttt{door} residual is smaller but less cleanly localized. The remaining
failure is consistent with handle/contact recovery sensitivity, and a
default-promotion candidate was not consistently better in spot checks. Since
the evidence did not show a focused target improvement with protected
non-regression, the loop did not accept a global policy variant. We therefore
leave this as an under-localized recovery/contact case rather than claiming it
is solved by high-level planning.

More generally, the RoboSuite failures show why recovery rules are not accepted
merely because they fire. Some monitors and recoveries helped earlier transport
and stack windows, but later low-level retries could add actions, mask the real
failure owner, or improve residual health while the environment success
predicate remained false. In Kintsugi, such candidates remain rejected hypotheses
until they can be expressed as KB-owned monitors with explicit trigger
conditions, repair targets, and regression evidence.

The same boundary appears in the more complex RoboCasa-style diagnostic runs,
which we do not include as a headline quantitative claim. The standing
OpenDoor/SingleCabinet scout plateaued at \(0\)--\(2/20\) after 38 edit
attempts because the dominant failure was no longer which symbolic subgoal to
pursue; it was the motor interface's ability to realize a narrow manipulation
effect under the available controller settings. In that case, edits could
identify the desired open-door relation and propose handle/contact recoveries,
but performance remained capped by controller gain, output limit, and OSC
servo bandwidth. This is a useful negative case for the method: once the
bottleneck is a motor ceiling rather than a missing symbolic contract,
additional KB edits should not be counted as learned policy unless they pass
the same environment success and non-regression gates.

\begin{wraptable}{r}{0.58\textwidth}
\vspace{-8pt}
\caption{\textbf{MetaWorld MT-50 non-ceiling tasks.}}
\label{tab:app-mt50}
\centering
\setlength{\tabcolsep}{3pt}
\renewcommand{\arraystretch}{1.04}
\begin{tabular*}{0.58\textwidth}{@{\extracolsep{\fill}}lcc@{}}
\toprule
\textbf{Task} & \textbf{Train} & \textbf{Held-out}\\
\midrule
basketball-v3 & 49/50 & 49/50\\
button-press-wall-v3 & 43/50 & 46/50\\
disassemble-v3 & 48/50 & 45/50\\
door-lock-v3 & 50/50 & 49/50\\
faucet-close-v3 & 50/50 & 49/50\\
peg-insert-side-v3 & 49/50 & 47/50\\
peg-unplug-side-v3 & 48/50 & 50/50\\
pick-place-v3 & 50/50 & 49/50\\
pick-place-wall-v3 & 49/50 & 50/50\\
push-back-v3 & 45/50 & 43/50\\
soccer-v3 & 43/50 & 45/50\\
stick-pull-v3 & 49/50 & 46/50\\
sweep-into-v3 & 43/50 & 40/50\\
\bottomrule
\end{tabular*}
\vspace{-8pt}
\end{wraptable}

For MetaWorld, the small train-to-held-out drop suggests that most of the
learned structure is not memorizing reset seeds. The remaining failures
concentrate in task families that require precise phase transitions or contact
timing, which is consistent with the method's boundary: Kintsugi can select and
repair symbolic schemas, but the continuous substrate must still provide
reliable primitive execution. Table~\ref{tab:app-mt50} lists only the non-ceiling
tasks; omitted MT-50 tasks are solved at \(50/50\) in the corresponding split.
The train split has 40/50 ceiling tasks and 2466/2500 successes; the held-out
split has 39/50 ceiling tasks and 2458/2500 successes.

\paragraph{Predicators / predictor-style details.}
The Predicators evaluation uses four task families with 10 test seeds and 50
tasks per family per seed. The suite result is \(1997/2000 = 99.85\%\), with
all residual misses isolated to the embodied PyBullet cover setting. The
\texttt{cover} domain is solved at \(500/500\) after the KB grounds the goal
block from the goal atom and uses cover-specific placement relations. The same
abstract cover rule transfers to \texttt{pybullet\_cover}, which reaches
\(497/500\) under robot-specific bindings; because the abstract setting is
already solved, these residual misses are better explained by Panda
IK/controller variance than by a missing symbolic rule. The \texttt{blocks}
domain reaches \(500/500\) after the policy owns table-height support facts,
goal-forest decomposition, y-axis table-clear tests, and strict clearance
margins. The \texttt{painting} domain reaches \(500/500\) after the KB makes
three hidden process constraints explicit: receptacle-specific grasping,
washing before drying, and opening the box lid before placement. These cases
show the same policy shift as in the text domains: once the missing condition
is represented as an executable symbolic contract, the executor no longer has
to rediscover it through search or prompt reasoning.

For RoboSuite, the result should be read as object-centric failure-localization
evidence rather than a full visual manipulation benchmark. The experiments use
simulator state, and some dynamic grounding and low-level execution still live
in task-local adapters. The useful signal is that failures can be localized
into grounding, operator, policy-schema, monitor, recovery, and planning layers,
and accepted patches can be distilled into the central KB.

\paragraph{RoboSuite protocol fidelity.}
The RoboSuite rows follow the same fixed-artifact evaluation contract as the
rest of the paper, but with an explicitly narrower substrate. Each protocol
round records the active KB, rollout bank, annotated trace, frontier analysis,
proposal, and status summary. The recorded protocol status marks the setting as
\emph{transitional}: \(S0\) is object-centric RoboSuite simulator state,
\(L0\) grounding is partly executed by task-local Python grounders, the runtime
brain is \texttt{none}, and the action owner is the symbolic executor. Learned
policy knowledge is counted only when it is represented as a KB-owned or typed
schema update and then checked by focused validation plus regression. Python
adapters are treated as execution plumbing or candidate prototypes, not as
standalone learned policy. Fixed seed windows are used only for paired
before/after localization; broader family or fresh-seed windows are used to
check regression.

\paragraph{Evidence for the RoboSuite boundary.}
The available evidence does not support the stronger claim that every remaining
RoboSuite drop has a single cause. It supports a narrower and more useful
claim: many drops are not explained by missing high-level task intent alone.
The traces and accepted/rejected edits localize them to the
symbolic-to-continuous execution boundary: object/contact grounding,
scene-conditioned sequencing, monitors and recoveries, and skill/action
bindings. The evidence is strongest when a focused before/after window improves
and a protected regression window does not degrade; diagnostic rows are used
only to localize the remaining boundary, not to claim solved behavior.

For tall-box transport, the accepted
\texttt{box\_transport\_yaw\_grounding\_v1} edit changes the Cereal window from
\(3/5\) to \(5/5\), while the PickPlace-family regression reaches \(29/30\).
This localizes the original failure to grounding and policy-schema binding for
yaw-aware side grasping rather than to a generic retry rule.

For nut tasks, the accepted edits form a more layered chain.
\texttt{round\_contact\_v1} moves RoundNut from \(0/5\) to \(5/5\);
\texttt{contact\_v2\_general} keeps RoundNut at \(5/5\) and moves SquareNut
from \(3/5\) to \(5/5\). Under strict environment success,
\texttt{square\_handle\_first\_approach\_v1} moves NutAssembly from \(2/5\)
to \(4/5\) with a \(19/20\) nut-family regression check. These edits support a
contact-frame, handle-grounding, and insertion-schema interpretation. Center
targeting and looser contact windows were not accepted when they worsened
focused or multi-nut validation.

For multi-nut composition, the evidence shifts from primitive execution to
planning-schema structure. The \texttt{multi\_nut\_insert} ordering principle
improves one NutAssembly window from \(0/5\) to \(3/5\). Later
scene-conditioned ordering reaches \(5/5\) on seeds 367247--367251 while
preserving the older \(4/5\) regression window. A fixed SquareNut-first order
solves one seed but breaks another, which is why the accepted repair is a
scene-conditioned subgoal-ordering contract rather than another primitive
threshold.

Recovery evidence is mixed in a useful way. Early transport iterations add
\texttt{payload\_lost}, release-event latching, and Stack soft-lift recovery,
moving the transport aggregate to \(64/70\). Later recovery candidates are
rejected when they only fire without improving final success, slow a successful
trajectory, or add a loop that is superseded by a shorter higher-level adaptive
schema. This is why we treat recovery as load-bearing but require KB-owned
monitor contracts with validation gates.

Finally, \texttt{tool\_hang} remains diagnostic rather than solved evidence.
Frame-grip and wall-insert probes improve intermediate relations such as
\texttt{frame\_tip\_close\_to\_base} and sometimes place
\texttt{frame\_tip\_to\_base} near the insertion window, but environment
success remains false and \texttt{frame\_rod\_between\_walls} remains false.
The boundary is not only task order; it is the missing frame-pivot /
stand-wall insertion schema with contact-preserving translation, orientation
channels, and explicit success monitors.

\FloatBarrier

\section{Typed KB Layers Are Load-Bearing Contracts}
\label{app:rq5-details}

This section expands Section~\ref{sec:rq5-ablation}. It asks whether the
\(S0/L0\)--\(L7\) structure is only descriptive or whether it defines
load-bearing execution contracts, making the frozen-artifact intervention
contract explicit. The intervention is performed on the frozen accepted ALFWorld
KB with no retraining and no test-time LLM calls.
Because different layers define different contracts, the diagnostic signal may
appear in execution, KB-query checks, planning, or a targeted probe rather than
in the same scalar metric for every row. In particular, the planning readout measures
task-schema and decomposition coverage; it is not identical to full
environment-valid execution.

\begin{table}[H]
\caption{\textbf{RQ5 layer-wise KB ablation on ALFWorld.}}
\label{tab:appendix-layer-ablation}
\centering
\setlength{\tabcolsep}{4pt}
\renewcommand{\arraystretch}{1.08}
\begin{tabularx}{\linewidth}{@{}L{0.16\linewidth}L{0.36\linewidth}ccc@{}}
\toprule
\textbf{Contract} &
\textbf{Intervention} &
\textbf{Exec.} &
\textbf{Query} &
\textbf{Plan}\\
\midrule
Full KB &
No removal &
\textbf{120/120} &
\textbf{2787/2787} &
\textbf{255/255}\\
S0 Source &
Cut off the goal/source record &
0/120 &
-- &
--\\
L0 Grounding &
Disable observation-to-state updates &
0/120 &
-- &
--\\
L1 Predicates &
Set all non-\texttt{always} predicates to false &
0/120 &
-- &
--\\
L2 Ontology &
Remove ontology, affordances, categories, and spatial priors &
0/120 &
2232/2787 &
255/255\\
L3 Operators &
Remove symbolic operators and effects &
120/120 &
2787/2787 &
255/255\\
L4 Policy &
Remove policy rules except the \texttt{LOOK} fallback &
0/120 &
2787/2787 &
255/255\\
L5 Recovery &
Disable invalid-command recovery &
120/120 &
2787/2787 &
255/255\\
L6 Experience &
Compare cold-start scaffold with accepted KB &
-- &
-- &
--\\
L7 Goals &
Remove task schemas and goal templates &
40/120 &
2130/2787 &
0/255\\
\bottomrule
\end{tabularx}
\end{table}
\begin{wraptable}{r}{0.56\textwidth}
\vspace{-8pt}
\caption{\textbf{Intra-layer ALFWorld subcomponent ablation.}}
\label{tab:component-ablation}
\centering
\setlength{\tabcolsep}{3pt}
\renewcommand{\arraystretch}{1.08}
\begin{tabular*}{0.56\textwidth}{@{\extracolsep{\fill}}lrrr@{}}
\toprule
\textbf{Variant} & \textbf{Execute} & \textbf{Query} & \textbf{Plan}\\
\midrule
Full KB              & 48/48 & 2787/2787 & 255/255\\
\(-\)spatial priors  & 48/48 & 2532/2787 & 255/255\\
Simple schema only   & 24/48 & 2160/2787 & 30/255\\
\(-\)round-robin rule & 46/48 & 2787/2787 & 255/255\\
\(-\)affordances     & 23/48 & 2487/2787 & 255/255\\
\bottomrule
\end{tabular*}
\vspace{-8pt}
\end{wraptable}

Table~\ref{tab:appendix-layer-ablation} supports RQ5 by showing layer-specific
failure signatures rather than uniform degradation. S0/L0/L1 form the execution
interface from source to grounded state to named predicates. L2 provides world
semantics and action applicability. L4 controls action selection. L7 controls
goal templates and task decomposition. L3, L5, and L6 are not fully diagnosed by
clean rollout success alone, so their roles are measured through targeted
probes for causal effects, recovery behavior, and learning provenance:
removing L3 effects keeps nominal execution at \(120/120\) in this runner but
drops the action-effect probe from \(11/11\) to \(0/11\); disabling L5 recovery
keeps clean rollouts intact but drops the invalid-command stress probe from
\(1/1\) to \(0/1\); and L6 is measured by the cold-start ledger moving from
\(0/134\) to \(134/134\) through accepted edit rounds.

The subcomponent ablations in Table~\ref{tab:component-ablation} refine the
layer-wise RQ5 result but are not the primary evidence for the typed-layer
design. Removing affordances damages execution because the executor loses
action-applicability constraints, while removing spatial priors is less harmful
on this particular ALFWorld diagnostic suite. Removing the round-robin rule
causes a smaller selective execution drop. The simple-schema variant compresses
multiple policy and task-schema structures, reducing both execution and
planning. These patterns are consistent with the broader S0/L0--L7 intervention
study.

\FloatBarrier

\section{Limitations and Future Work}
\label{app:limitations-future}

\paragraph{Limitations.}
Kintsugi is a white-box task-knowledge layer, not an end-to-end robot-learning
system. The strongest results assume that \(S0\) already exposes a reliable
state interface: TextWorld-style observations and admissible actions for
ALFWorld, public text observations for WebShop, recipe/inventory state for
TextCraft, simulator state for RoboSuite, and object-centric state for
MetaWorld and Predicators. This is the right abstraction for testing whether
typed policy knowledge can be learned, edited, verified, and executed, but it
does not solve perception. When an object is misdetected, a relation is
estimated incorrectly, or a visual tracker loses an object, the current
experiments assume that this error has already been handled below \(S0\).

The second limitation is the skill and controller boundary. The KB can choose
a symbolic subgoal, bind a skill, and monitor whether the intended relation
becomes true, but it does not learn arbitrary low-level motor control from
scratch. This is why RoboSuite is reported as boundary evidence and why
\texttt{tool\_hang} remains unresolved in the single-arm aggregate. The
symbolic policy can identify that frame insertion and hook placement require
contact-preserving motion and orientation control, but the current action
interface does not provide a verified primitive that keeps the frame inside the
narrow insertion window while placing the tool. The RoboCasa-style
OpenDoor/SingleCabinet diagnostic shows the same limitation more sharply:
after many edit attempts, the bottleneck is controller gain, output range, and
OSC servo bandwidth rather than another missing high-level symbolic rule.

The third limitation is verifier coverage. A typed edit is only as safe as the
focused and protected banks used to accept it. The verifier prevents many
obvious forms of drift, but it cannot certify behavior outside the covered
state distribution. This is especially important for WebShop and robot
manipulation. In WebShop, public snippets and search ranking can hide the
decisive product evidence, so a local symbolic constraint rule may improve one
window while failing another. In manipulation, a recovery rule can improve an
intermediate residual while still failing the task success predicate or
slowing a previously successful trajectory.

Finally, Kintsugi's current evidence is about task-level policy knowledge, not
about learning the entire robot stack. The framework assumes a substrate that
can expose objects, relations, actions, or calibrated skills with enough
stability for symbolic contracts to be meaningful. This is a deliberate scope
choice: it isolates whether task knowledge can be stored as an editable,
verifier-gated artifact. It also defines the next research problem.

\paragraph{Broader impacts.}
The intended positive impact is more maintainable embodied-agent behavior:
policy knowledge is inspectable, edits are local, and deployed action selection
does not require a test-time LLM. This can reduce opaque prompt drift and make
failure analysis easier for safety reviews, debugging, and reproducibility.
The corresponding risk is over-trusting an executable artifact outside its
verified state, skill, or perception contract. A KB that is correct in a
simulator or text environment may become unsafe if deployed with an unreliable
vision system, an uncalibrated controller, or an incomplete protected
regression bank. Responsible use therefore requires benchmark-scoped claims,
human review before real-world deployment, explicit \(S0\) and skill-interface
contracts, verifier-gated updates, and regression banks that match the intended
deployment distribution.

\paragraph{Future work: replacing oracle \(S0\) with vision.}
The central next step is to replace oracle or simulator object-centric \(S0\)
with a vision-derived state interface while preserving the same KB and
verification discipline. The goal is not to discard the symbolic artifact, but
to make \(S0\) a learned perception and state-estimation contract. A visual
\(S0\) should provide object slots or tracks, categories, poses, attributes,
relations, uncertainty, and temporal persistence. The existing \(L0\) grounding
layer would then map these perceptual entities into KB predicates such as
\texttt{held}, \texttt{open}, \texttt{inside}, \texttt{near}, \texttt{aligned},
or task-specific contact relations.

This vision extension should be staged rather than treated as a single
end-to-end black box. First, train or plug in a perception module that produces
object-centric state with calibrated uncertainty. Second, add grounding checks
that compare visual predicates against environment or human-labeled evidence,
so failures can be localized to perception, grounding, symbolic policy, or
skill execution. Third, let Kintsugi edit \(L0\)--\(L7\) while keeping \(S0\)
versioned: if a failure is caused by missed detection or relation estimation,
the system should update the visual grounding contract or request additional
perceptual supervision rather than incorrectly adding a policy rule. Fourth,
extend the verifier to include visual regression banks, so a new grounding rule
or detector threshold must improve the target visual failure without breaking
previous object/relation recognition.

The longer-term direction is an end-to-end embodied system in which perception,
state estimation, symbolic task knowledge, and skills are all trainable but
remain modularly inspectable. The policy should still execute through an
artifact: visual \(S0\) produces uncertain object-centric state, \(L0\) grounds
that state into predicates, \(L1\)--\(L7\) select and monitor symbolic actions,
and the skill layer executes calibrated motor commands. The LLM can remain a
proposal engine for edits and diagnostics, but deployed action selection should
continue to be owned by the executable artifact. This preserves the main
benefit of Kintsugi while moving from oracle object-centric representations to
vision-based, end-to-end embodied learning.

\FloatBarrier